\documentclass[lettersize,journal]{IEEEtran}
\usepackage{amsmath,amsfonts}
\usepackage[ruled,linesnumbered,vlined]{algorithm2e}
\usepackage{array}
\usepackage[caption=false,font=normalsize,labelfont=sf,textfont=sf]{subfig}
\usepackage{textcomp}
\usepackage{stfloats}
\usepackage{url}
\usepackage{verbatim}
\usepackage{graphicx}
\usepackage{cite}
\usepackage{algpseudocode}  
\usepackage{multirow}
\usepackage{booktabs}
\usepackage{xcolor}
\usepackage{tikz}

\usepackage[hidelinks,
            colorlinks=False,
            linkcolor=black,       
            anchorcolor=black,  
            citecolor=black,       
            ]{hyperref}
            
\definecolor{lime}{HTML}{A6CE39}
\DeclareRobustCommand{\orcidicon}{%
	\begin{tikzpicture}
	\draw[lime, fill=lime] (0,0) 
	circle [radius=0.16] 
	node[white] {{\fontfamily{qag}\selectfont \tiny ID}};
	\draw[white, fill=white] (-0.0625,0.095) 
	circle [radius=0.007];
	\end{tikzpicture}
	\hspace{-2mm}
}

\foreach \x in {A, ..., Z}{%
	\expandafter\xdef\csname orcid\x\endcsname{\noexpand\href{https://orcid.org/\csname orcidauthor\x\endcsname}{\noexpand\orcidicon}}
}

\begin{document}

\title{TMUAD: Enhancing Logical Capabilities in Unified Anomaly Detection Models with a Text Memory Bank}

\author{Jiawei Liu\orcidA{}, \IEEEmembership{Member,~IEEE, }
Jiahe Hou\orcidB{}, 
Wei Wang\orcidC{}, 
Jinsong Du\orcidD{}, 
Yang Cong\orcidE{}, \IEEEmembership{Senior Member,~IEEE, }
and Huijie Fan\orcidF{}, \IEEEmembership{Member,~IEEE}
\thanks{This work is supported by the National Natural Science Foundation of China (62273339, U24A201397), the LiaoNing Revitalization Talents Program (XLYC2403128), and the Natural Science Foundation of Liaoning Province (2024-MSBA-86).}
\thanks{Wei Wang and Huijie Fan are the corresponding authors of this work.}
\thanks{
Jiawei Liu, Wei Wang, and Jinsong Du are with the Shenyang Institute of Automation, Chinese Academy of Sciences, Shenyang 110016, China, also with  the Liaoning Liaohe Laboratory, Shenyang 110016, China, and also with the Key Laboratory on Intelligent Detection and Equipment Technology, Shenyang 110169, China (e-mail: liujiawei@sia.cn; wangwei2@sia.cn; jsdu@sia.cn).}

\thanks{Jiahe Hou is with the Shenyang Institute of Automation, Chinese Academy of Sciences, Shenyang 110016, China, also with the Liaoning Liaohe Laboratory, Shenyang 110016, China, also with the Key Laboratory on Intelligent Detection and Equipment Technology, Shenyang 110169, China, and also with the University of Chinese Academy of Sciences, Beijing, 100049, China (e-mail: houjiahe25@mails.ucas.ac.cn).}

\thanks{Huijie Fan is with the State Key Laboratory of Robotics and Intelligent Systems, Shenyang Institute of Automation, Chinese Academy of Sciences, Shenyang, 110016, China (e-mail: fanhuiie@sia.cn).}

\thanks{Yang Cong is with the College of Automation Science and Engineering, South China University of Technology, Guangzhou, 510640, China (e-mail: congyang81@gmail.com).}

}


\maketitle

\begin{abstract}
Anomaly detection, which aims to identify anomalies deviating from normal patterns, is challenging due to the limited amount of normal data available. Unlike most existing unified methods that rely on carefully designed image feature extractors and memory banks to capture logical relationships between objects, we introduce a text memory bank to enhance the detection of logical anomalies.  Specifically, we propose a Three-Memory framework for Unified structural and logical Anomaly Detection (TMUAD). First, we build a class-level text memory bank for logical anomaly detection by the proposed logic-aware text extractor, which can capture rich logical descriptions of objects from input images. Second, we construct an object-level image memory bank that preserves complete object contours by extracting features from segmented objects. Third, we employ visual encoders to extract patch-level image features for constructing a patch-level memory bank for structural anomaly detection. These three complementary memory banks are used to retrieve and compare normal images that are most similar to the query image, compute anomaly scores at multiple levels, and fuse them into a final anomaly score. By unifying structural and logical anomaly detection through collaborative memory banks, TMUAD achieves state-of-the-art performance across seven publicly available datasets involving industrial and medical domains.
The model and code are available at \url{https://github.com/SIA-IDE/TMUAD}.

\end{abstract}

\begin{IEEEkeywords}
Anomaly detection, logical anomaly, structural anomaly, memory bank, text descriptions, unified model, segmentation.
\end{IEEEkeywords}

\section{Introduction}

\IEEEPARstart{I}{ndustrial} anomaly detection is critical for quality inspection and predictive maintenance, involving the identification of deviations from normal patterns with limited or no prior knowledge of potential anomalies \cite{golan2018deep,roth2022towards,10657290}. Industrial anomaly images can be roughly categorized into two types: structural anomalies \cite{bergmann2019mvtec, zou2022spot} and logical anomalies \cite{bergmann2022beyond}. 
Structural anomaly detection methods include reconstruction-based methods (autoencoder \cite{beizaee2025correcting, wang2021cognitive, hou2021divide,cai2024rethinking} and diffusion \cite{zhang2024realnet, he2024diffusion, liu2024residual, yao2024glad}), teacher-student architecture-based methods \cite{deng2022anomaly, liu2024dual ,10448432, WU2024159}, and memory bank-based methods \cite{roth2022towards, xie2023pushing, bae2023pni, 10716437,10378462,hyun2024reconpatch,10446766}. These methods 1) typically focus on detecting local regions and surfaces, such as scratches, dents, stains, cracks, or contaminants in manufactured products \cite{zhang2025towards}, and 2) fail in logical anomaly detection, as such anomalies may appear normal in local details but manifest as object missing, incorrect position, or incorrect combinations of normal elements, requiring the capture of long-range and global dependencies \cite{10943358, 10483687, 10484326, kim2024few, gu2025univad}.

\begin{figure}[t]
\centering
\includegraphics[width=\columnwidth]{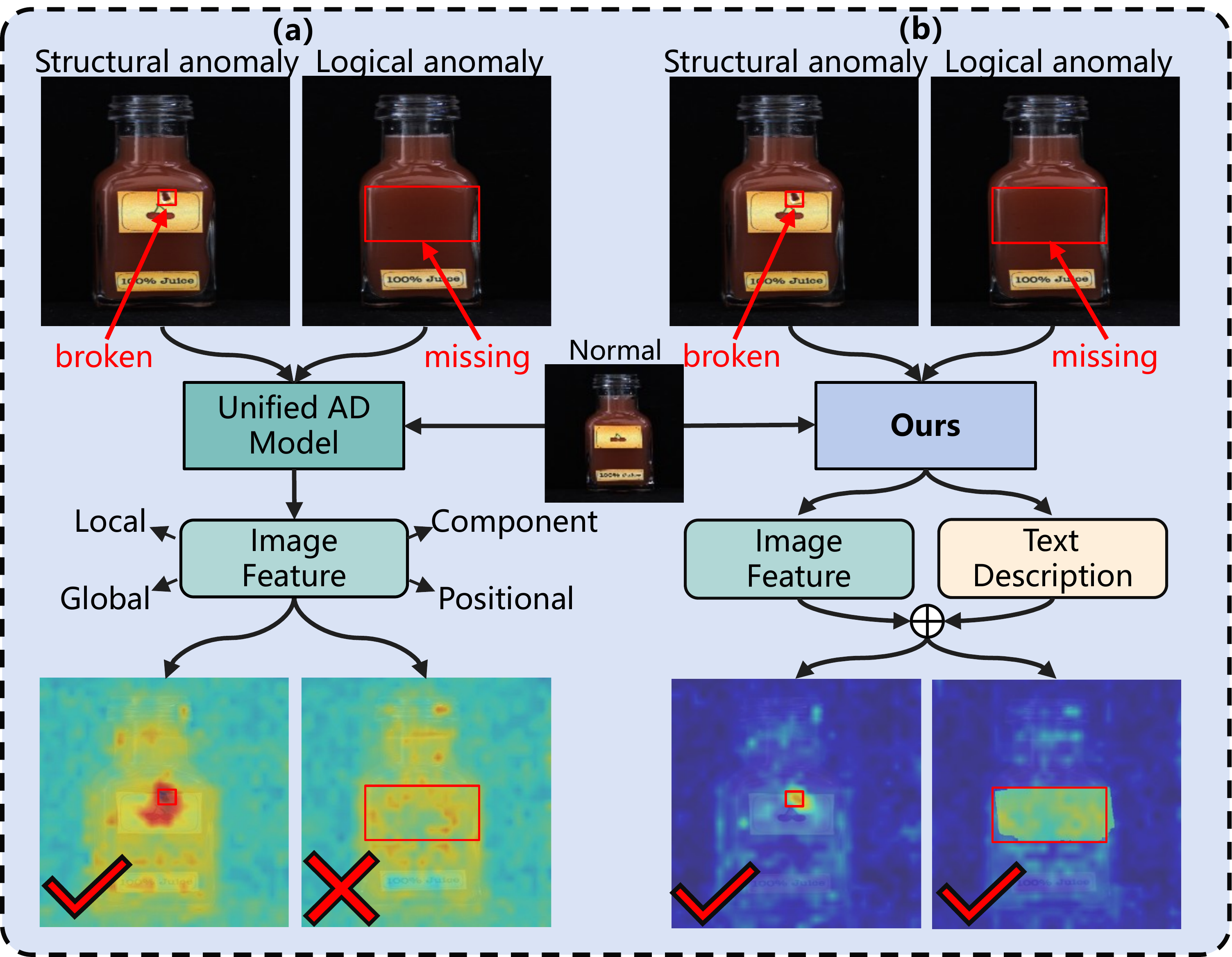} 
\caption{
Comparison between our TMUAD and existing frameworks.
(a) The unified anomaly detection model in prior work \cite{gu2025univad} extracts image features to jointly address logical and structural anomalies.
(b) In contrast, our TMUAD introduces text descriptions to improve logical anomaly detection, while using image features for structural anomaly detection.
}
\label{introduction}
\end{figure}

To unify structural anomaly and logical anomaly detection, current memory-based anomaly detection methods \cite{peng2025sam, kim2024few, gu2025univad, zhang2025towards} determine the presence of anomalies (image-level anomalies) and detect anomalous regions (pixel-level anomalies) by extracting image features. As shown in Fig. \ref{introduction}(a), these methods aim to simultaneously extract global and local image features for detecting structural anomalies, as well as component and positional image features for identifying logical anomalies. Currently, techniques for extracting global and local features are similar, typically employing Vision Transformers to encode images into patch features. Recent approaches generally enhance performance by designing new strategies for extracting components and positional features to understand the relationships between components. For example, GCAD \cite{bergmann2022beyond} effectively emphasizes causal relationships in images by extracting causal matrices from patch features and applying causality graph sparsification. UniVAD \cite{gu2025univad} uses a graph-enhanced component modeling, which abstracts inter-component relationships into graph structures. Logical relationships are captured through nodes and edges, enabling the extraction of relational features. SAM-LAD \cite{peng2025sam} initially employs a base model to perform zero-shot segmentation and subsequently models inter-component relationships by constructing a similarity matrix to extract logical dependencies.
However, there are two issues in the anomaly detection process: 1) Both structural anomaly and logical anomaly detection rely on searching and comparing image features stored in the memory bank with those of the test image. 2) High-dimensional image features are often extracted to improve logical anomaly detection capabilities. This leads to poor performance in logical anomalies because most image feature channels are unrelated to logical anomalies.

In this paper, instead of an elaborately designed image feature memory bank, we introduce a text memory bank to enhance the detection of logical anomalies, as shown in Fig. \ref{introduction}(b). Text, as the primary medium of natural language, is a expression form rich in logical information, effectively encoding object categories, counts, locations, sizes, and combination relationships.
Compared to images, text can present information in a more concise and structured manner, filtering out a large amount of redundant content unrelated to logic, which is crucial for improving the performance of logical anomaly detection.

Specifically, we propose a Three-Memory framework for Unified structural and logical Anomaly Detection (TMUAD), consisting of class-level text memory bank, object-level image memory bank, and patch-level image memory bank.
We first propose a logic-aware text extractor to capture rich logical descriptions of all objects from input images, containing object categories, counts, locations, sizes, and combination relationships.
These text descriptions compose a text memory bank for logical anomaly detection.
We then construct an object-level image memory bank by extracting image features of segmented objects to retain complete object contours. Finally, we use visual encoders to extract patch-level image features for structural anomaly detection.
These three complementary memory banks are used to retrieve and compare normal images that are most similar to the query image, compute individual anomaly scores, and fuse them into a final anomaly detection score. 
Our framework unifies structural anomalies and logical anomalies using three collaborative memory banks, enabling  more accurate logical anomalies detection in object counts and locations through explicit text descriptions.
The contributions of this paper are summarized as follows:
\begin{enumerate}
\item We propose a unified structural and logical anomaly detection framework, consisting of class-level text, object-level image, and patch-level image memory bank. Our class-level text memory bank can significantly enhance the logical anomaly detection performance.
\item We propose a text-based anomaly detection pipeline, including a plug-and-play logic-aware text extractor to capture rich logical information between objects for logical anomaly detection, and a compatible anomaly score calculation method for text by reverse-locating images from anomaly-related text. 
\item Our unified anomaly detection framework achieves state-of-the-art performance in logical and structural anomaly detection across seven publicly available datasets involving industrial and medical domains.
\end{enumerate}

\section{Related work}
\subsection{Memory Bank-based Anomaly Detection}
Since anomaly samples are very scarce in real-world environments, many methods focus on achieving unsupervised anomaly detection by constructing memory banks. PatchCore \cite{roth2022towards} uses patch features extracted by ImageNet pre-trained encoders to construct memory bank, and calculate the similarity between query samples and memory bank features to detect anomalies. PMB \cite{10018373} combines partition memory bank Module to build storage templates for each channel to retain detailed features. PBAS \cite{10716437} uses a feature projector layer based on PatchCore \cite{roth2022towards} to improve anomaly detection accuracy. OPFA \cite{zhou2024outlier} combines GMM \cite{reynolds2015gaussian} to model the distribution of memory bank data, thereby identifying abnormal data more effectively. These models perform very well on structural anomaly, but cannot effectively distinguish logical anomaly.

Along with the first logical anomaly dataset MVTec LOCO \cite{bergmann2022beyond}, many methods began to focus on a unified framework for detecting structural anomaly and logical anomalies. GCAD \cite{bergmann2022beyond} removes noise from causal graphs through simple and effective causal graph sparsification, thereby improving the interpretability of causal relationships in images. UniVAD \cite{gu2025univad} constructs structured combinatorial relationships between objects by combining component segmentation techniques and graph-enhanced component model to achieve effective determination of combinatorial logic rules. These methods achieve the extraction of logical information effectively by designing specialized components, but the features lack interpretability and have a large amount of redundant information, resulting in limited performance of logical anomaly detection.

\subsection{Vision Language Models–based Anomaly Detection.}

In recent years, Vision Language Models (VLMs)\cite{gu2024anomalygpt, 10819451,ma2023crepe , ma2025aa, zhang2025towards, mitra2024compositional, kwon2025logicqa,zeng2024investigating} have provided new technological tools for anomaly detection due to their revolutionary reasoning capabilities in various vision tasks. SAM \cite{kirillov2023segment} has a powerful zero-sample image segmentation capability, which allows us to segment unseen objects without training. CLIP \cite{radford2021learning} uses contrastive learning to map images and texts to a unified vector space, supporting tasks such as image classification and retrieval. Qwen2-VL \cite{wang2024qwen2} possess advanced question-answering and fine-grained localization capabilities, offering new technical approaches for anomaly detection tasks.

AnomalyGPT \cite{gu2024anomalygpt} freezes the main parameters of the language model and adds a prompt learner to it, which embeds image features and text features into the original input latent space of the visual language model, and achieves fine-tuning by training the parameters of the cue embedding network. FocusCLIP \cite{10819451} uses the anomaly capture module to refine these defect regions, improving its ability to identify detailed anomalies. AA-CLIP \cite{ma2025aa} achieves fine-tuning by introducing a residual adapter layer to the CLIP, giving the model the ability to distinguish between normal and abnormal. LogicQA \cite{kwon2025logicqa} guides the VLMs through the designed prompts to summarize the patterns of the samples and to detect anomalies. LogSAD \cite{zhang2025towards} employs GPT-4 to automatically induce rules and utilizes a matching mechanism to detect anomalies by comparing object types and verifying potential rule violations. Although this method incorporates textual information to assess the consistency of object categories, it still relies primarily on image features rather than textual descriptions when processing positional relationships, thereby limiting the model’s overall performance. Moreover, this approach lacks the ability to achieve pixel-level precise localization of detected text anomalies. To address these limitations, this paper proposes enhancing the extraction of textual information and introducing a text-query-based localization strategy to improve both detection accuracy and localization effectiveness.

\begin{figure*}[t]
\centerline{\includegraphics[width=\textwidth]{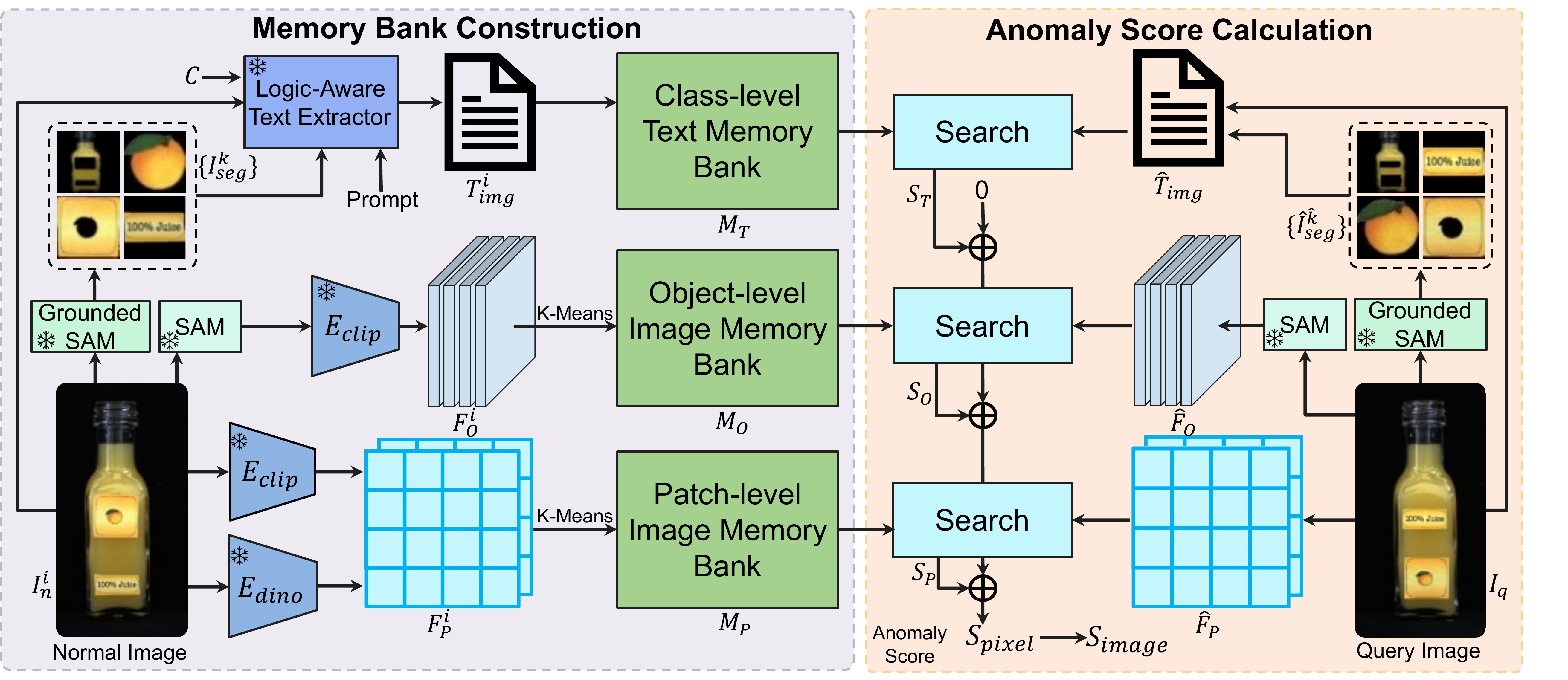}}
\caption{
The proposed Three-Memory framework for Unified Structural and logical Anomaly Detection (TMUAD). Our TMUAD incorporates three complementary memory banks: a class-level text memory bank ($M_T$), an object-level image memory bank ($M_O$), and a patch-level image memory bank ($M_P$). The left panel illustrates the construction of these memory banks, while the right panel depicts the process of anomaly score calculation.}
\label{main_framework_diagram}
\end{figure*}

\section{Method}
\subsection{Overview}

Our goal is to develop a multi-memory framework to unify logical and structural anomaly detection. For structural anomaly detection, current patch-based methods\cite{ roth2022towards, hyun2024reconpatch,gu2025univad,bergmann2022beyond} perform well but exhibit limitations in effectively detecting logical anomalies. To address these limitations, we propose three complementary strategies: 1) retain patch-level image features to construct a patch-level image memory bank $M_P$, thereby preserving the model’s capability to detect structural anomalies, 2) introduce a logic-aware textual description of images to significantly enhance logical anomalies detection of unified anomaly detection frameworks by constructing a class-level text memory bank $M_T$, and 3) introduce object-level image features from segmented images to build an object-level image memory bank $M_O$, which compensates for the inability of patch-level features to fully capture the shape and structure of individual objects.
As shown in Fig. \ref{main_framework_diagram}, Our TMUAD framework consists of memory bank construction in Sec.~\ref{section3_2} and anomaly score calculation in Sec.~\ref{section3_3}.

\subsection{Memory Bank Construction}\label{section3_2}
Our TMUAD takes a normal image $I_n^i$ as input and extracts corresponding text descriptions $T_{img}^i$, object-level image features $F_O^i$, and patch-level image features $F_P^i$ to construct three distinct memory banks: $M_T$, $M_O$, and $M_P$. The class-level text memory bank $M_T$ contains information such as object categories, counts, locations, and sizes. By capturing the attributes of each object category, $M_T$ effectively represents the logical relationships between object categories, making it suitable for detecting logical anomalies, including missing objects, incorrect object combinations, or abnormal spatial distributions.
The object-level image memory bank $M_O$ contains the visual features of individual objects, capturing their complete structural characteristics. The patch-level image memory bank $M_P$ stores multi-scale local visual features, enabling the representation of fine-grained details across the whole image.
In addition, we use the K-means algorithm to cluster the features of $M_O$ and $M_P$ to reduce storage space and improve retrieval efficiency \cite{gu2025univad}.

\subsubsection{Class-level Text Memory Bank} We construct a class-level text memory bank $M_T:=\{T_{img}^i\}_{i=1}^{n_i}$ from normal images $I_n^i$ in the training set, where $T_{img}^i$ represents text descriptions of $I_n^i$ for logical anomaly detection, $n_i$ is the total number of normal images. $T_{img}^i:=\{T_{class}^j\}_{j=1}^{n_j}$ comprises the textual descriptions of different object categories $T_{class}^j$ in $I_n^i$, where $T_{class}^j:=\{O_{class}^j,O_{num}^j,O_{pos}^j,O_{size}^j\}$ contains the object category $O_{class}^j$, number $O_{num}^j$, position $O_{pos}^j$, size $O_{size}^j$. $n_{j}$ is the number of object categories in $I_n^i$.

\begin{figure}[t]
\centering
\includegraphics[width=\columnwidth]{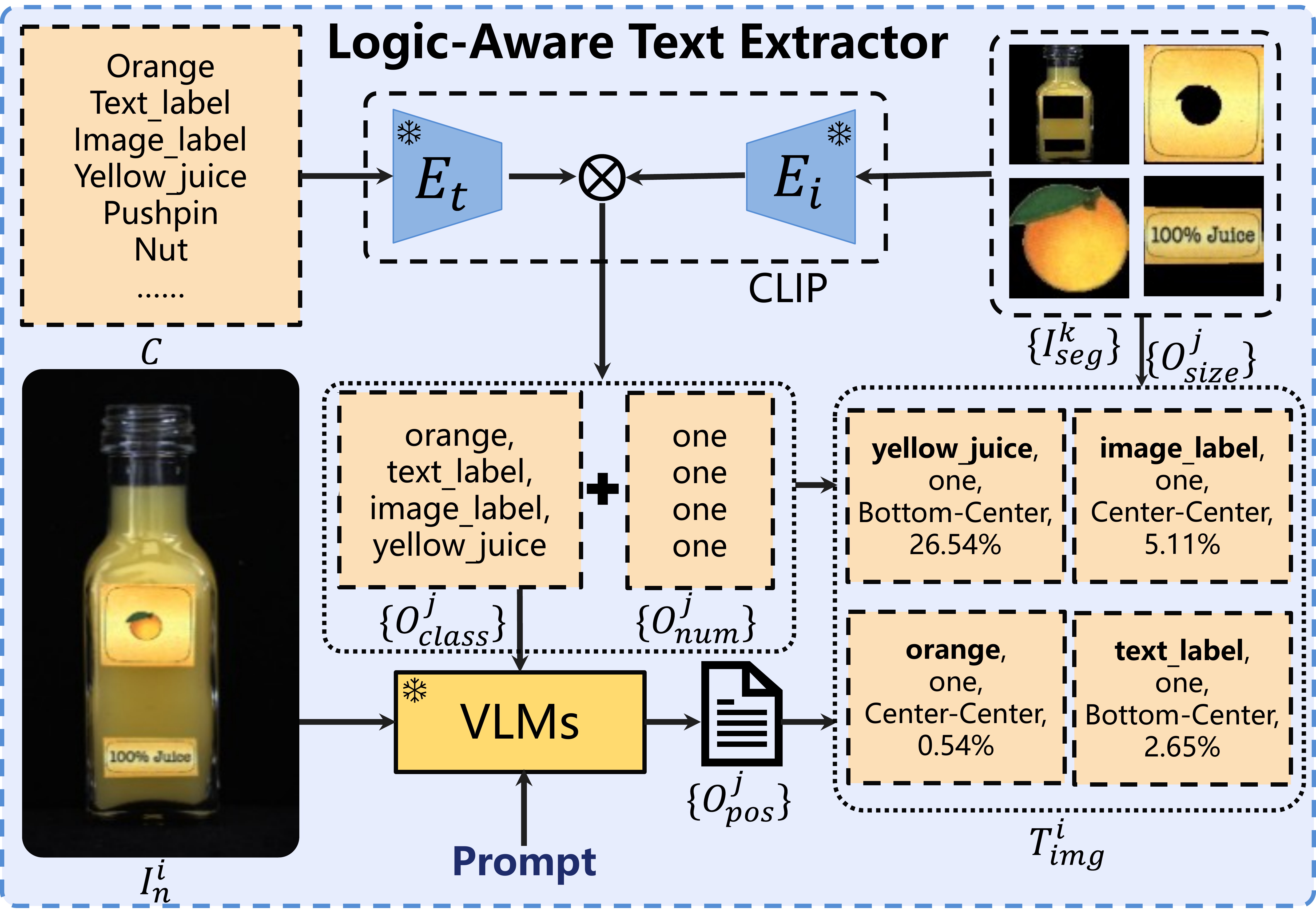} 
\caption{Proposed logic-aware text extractor for logical anomaly detection. Structured text descriptions rich in logical information can be extracted from normal images $I_n^i$, including the category $O_{class}^j$ and number $O_{num}^j$ of segmented objects obtained using CLIP \cite{radford2021learning}, fuzzy position $O_{pos}^j$ descriptions obtained using VLMs \cite{wang2024qwen2}, and the size of segmented objects $O_{size}^j$.}
\label{Visual_Describer}
\end{figure}

As shown in Fig. \ref{Visual_Describer} and Algorithm~\ref{Algorithm1}, we propose a logic-aware text extractor to capture rich logical information between objects for logical anomaly detection. 
1) To obtain object categories $O_{class}^j$, we first use Grounded SAM \cite{ren2024grounded} to segment a normal image $I_n^i$ into different objects $I_{seg}^k$, i.e., $I_{seg}^k=I_m^k\odot I_n^i$, $\{I_m^k\}=\mathrm{GSAM}(I_n^i)$, where $n_{k}$ is the number of segmented objects in the normal image $I_n^i$ by Grounded SAM \cite{ren2024grounded}. Then, we use CLIP \cite{radford2021learning} to obtain the categories $C_k$ of the segmented objects, i.e., $C_k=\mathrm{CLIP}(I_{seg}^k,C)$, where $C$ is a
list of predefined object categories.
2) To obtain object counts $O_{num}^j$, we merge the segmented images $I_{seg}^k$ of the same category and record their counts.
3) To obtain object position $O_{pos}^j$, we use VLM (i.e., Qwen2-VL \cite{wang2024qwen2}) to roughly describe the position of objects with prompts containing object categories, $\{O_{pos}^j\}= \mathrm{VLM}(\{O_{class}^j\}, I_{n}^i)$.
4) To obtain object size $O_{size}^j$, we calculate the area percentage of each segmented object relative to the whole normal image $I_n^i$, i.e., $\{O_{size}^j\} = \mathrm{Area}(\{I_{class}^j\})$, where $I_{class}^j$ is the merged mask of all objects belonging to category $O_{class}^j$.
Unlike employing VLMs to generate a single textual description for the entire image \cite{gu2024anomalygpt,11093352,hurst2024gpt,liu2023mitigating}, which is difficult to accurately describe the number, size, and combination of objects, we propose logic-aware text descriptions based on segmented images, facilitating logical anomaly detection such as incorrect object combinations, incorrect positions, and missing objects.

\begin{algorithm}[t]
   \SetKwData{Left}{left}\SetKwData{This}{this}\SetKwData{Up}{up}
   \SetKwFunction{CLIP}{CLIP}\SetKwFunction{return}{return}\SetKwFunction{VLM}{VLM}\SetKwFunction{GSAM}{GSAM}\SetKwFunction{Area}{Area}\SetKwFunction{Sort}{Sort}
   \SetKwInOut{Input}{Input}\SetKwInOut{Output}{Output}\SetKwFunction{Detach}{Detach}
   \Input{The $i$-th normal image ${I}_{n}^i$ in the training set. $n_i$ is the total number of normal images. A list of predefined object categories $C$.}
   \Output{A text memory bank $M_{T}$.}
   \BlankLine
   \For{$i\leftarrow 1$ \KwTo $n_i$}{
        \tcp{1) Obtain object categories.}
        $\{I_{m}^k\}, n_k\leftarrow \GSAM(I_n^i)$\\
        
        \For{$k\leftarrow 1$ \KwTo $n_k$}{
            $I_{seg}^k\leftarrow I_{m}^k\ \odot I_n^i$  \\
            $C_k\leftarrow \CLIP (I_{seg}^k,C)$
        }
        $\{C_k\},\{I_{seg}^k\} \leftarrow \Sort (\{C_k\},\{I_{seg}^k\})$, $j\leftarrow 0$\\
        \tcp{2) Obtain object counts.}
        $\{I_c|c\in C\}\leftarrow 0$\\
        \For{$k\leftarrow 1$ \KwTo $n_k$}{
            \uIf{$C_k$ \bf{not} \bf{in} $\{C_{k-1}\}$}{
                $j\leftarrow j+1$, $O_{class}^j \leftarrow C_k$\\
                $O_{num}^j\leftarrow 1$, $I_{class}^j \leftarrow 0$
            }
            \Else{
                $O_{num}^j\leftarrow O_{num}^j + 1$, $I_{class}^j \leftarrow I_{class}^j+I_{m}^k$
            }
            $I_{c}\leftarrow I_{c}+I_{m}^k$\Comment{$c=O_{class}^j $}
        }
        \tcp{3) Obtain object positions.}
        $n_j \leftarrow j$, $\{O_{pos}^j\}\leftarrow \VLM(\{O_{class}^j\}, I_{n}^i)$ \\
        \tcp{4) Obtain object sizes.}
        $\{O_{size}^j\} \leftarrow \Area(\{I_{class}^j\})$\\
         \For{$j\leftarrow 1$ \KwTo $n_j$}{
            $T_{class}^j\leftarrow \{O_{class}^j,O_{num}^j,O_{pos}^j,O_{size}^j\}$}
        $T_{img}^i\leftarrow \{T_{class}^j\}_{j=1}^{n_j}$ 
    }
    $M_{T} \leftarrow \left \{ T_{img}^i \right \} _{i=1}^{n_i}$
   \caption{Class-level Text Memory Bank.}\label{Algorithm1}
\end{algorithm}

\subsubsection{Object-level Image Memory Bank} We extract the visual features $F_{seg}^s$ of each segmented object $I_{seg}^s$ from the normal image $I_n^i$ by the pre-trained SAM \cite{kirillov2023segment} and the image encoder of CLIP \cite{radford2021learning}, i.e., $\{I_{seg}^s\}_{s=1}^{n_s}=\mathrm{SAM}(I_n^i)$, $F_{seg}^s = E_{clip}(I_{seg}^s)$, where $n_s$ is the number of segmented objects in the normal image $I_n^i$ by SAM \cite{kirillov2023segment}. The object-level image memory bank can be denoted as $M_o:=\ {\textstyle \bigcup_{i=1}^{n_i}}F_O^i$, where $F_O^i=\left \{ F_{seg}^s \right \} _{s=1}^{n_s}$ is the object-level image features of $I_n^i$. This memory bank offers several advantages. 1) By utilizing segmented objects, it effectively eliminates background interference since anomalies typically occur within the foreground objects of images. 2) In contrast to the class-level text memory bank or the patch-level image memory bank, the object-level image memory bank enables the retrieval and comparison of similar objects across different images.

\subsubsection{Patch-level Image Memory Bank} 
We employ CLIP \cite{radford2021learning} and DINOv2 \cite{oquab2023dinov2} as vision encoders to extract multi-scale visual features $F_P^i$ from normal images $I_n^i$, i.e., $F_P^i=\{E_{clip}(I_n^i),E_{dino}(I_n^i)\}$. 
These multi-scale features $F_P^i$ are obtained from multiple intermediate layers of the encoder: shallow layers primarily capture low-level visual cues such as edges, textures, and color, whereas deeper layers encode high-level semantic representations and global contextual information \cite{jiang2023clip}.
The patch-level image memory bank can be denoted as $M_P:=\left \{ F_P^i \right \}_{i=1}^{n_i}$.

\begin{figure}[t]
\centering
\includegraphics[width=\columnwidth]{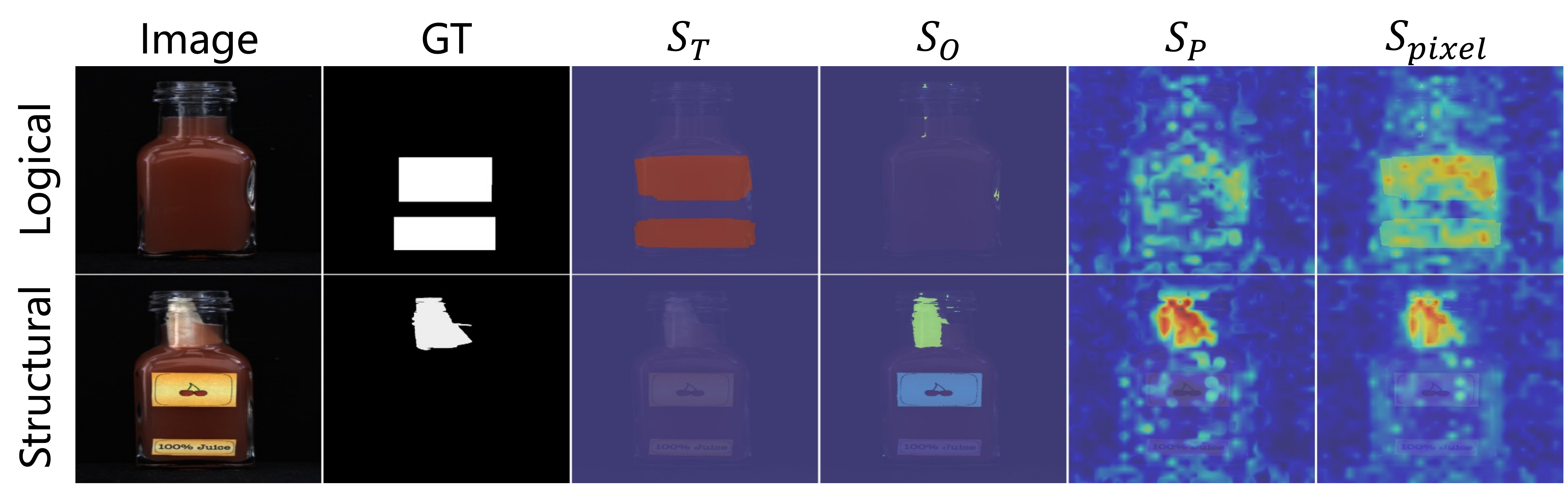} 
\caption{Fusion of anomaly scores from three different memory banks.}
\label{tensor_patch}
\end{figure}

\subsection{Anomaly Score Calculation}\label{section3_3}

Our TMUAD inputs a query image $I_q$, extracts corresponding text descriptions ($\hat{T}_{img}$), object-level image features $\hat{F}_O$, and patch-level image features $\hat{F}_P$, searches and compares the most similar features in the three memory banks, calculates three anomaly scores for each memory bank (class-level text anomaly score $S_T$, object-level image anomaly score $S_O$, and patch-level image anomaly score $S_P$), and finally fuses these anomaly scores for anomaly detection (i.e., ${S}_{pixel}$ in Fig. \ref{tensor_patch}).

\subsubsection{Class-level Text Anomaly Score}
Current anomaly detection methods \cite{ma2025aa,he2024diffusion,gu2024anomalygpt} typically compute pixel-level anomaly scores to localize anomalous regions, while image-level anomaly scores are used to determine the presence of anomalies.
The image-level anomaly scores can be derived from pixel-level scores, for example by taking their maximum or average value \cite{gu2025univad,beizaee2025correcting,10716437}.
However, when anomaly detection is performed with text, only image-level anomaly scores are readily available, whereas pixel-level scores are challenging to obtain \cite{11093352,zhang2025towards,kwon2025logicqa}.
To address this limitation, we propose a pixel-level calculation method for class-level text anomaly score by reverse-locating images from anomaly-related text. This method is compatible with widely adopted anomaly score calculation pipelines, enabling the computation of both image-level and pixel-level anomaly scores using a text-based memory bank.

\begin{figure}[h]
   \centering
   \includegraphics[width=\columnwidth]{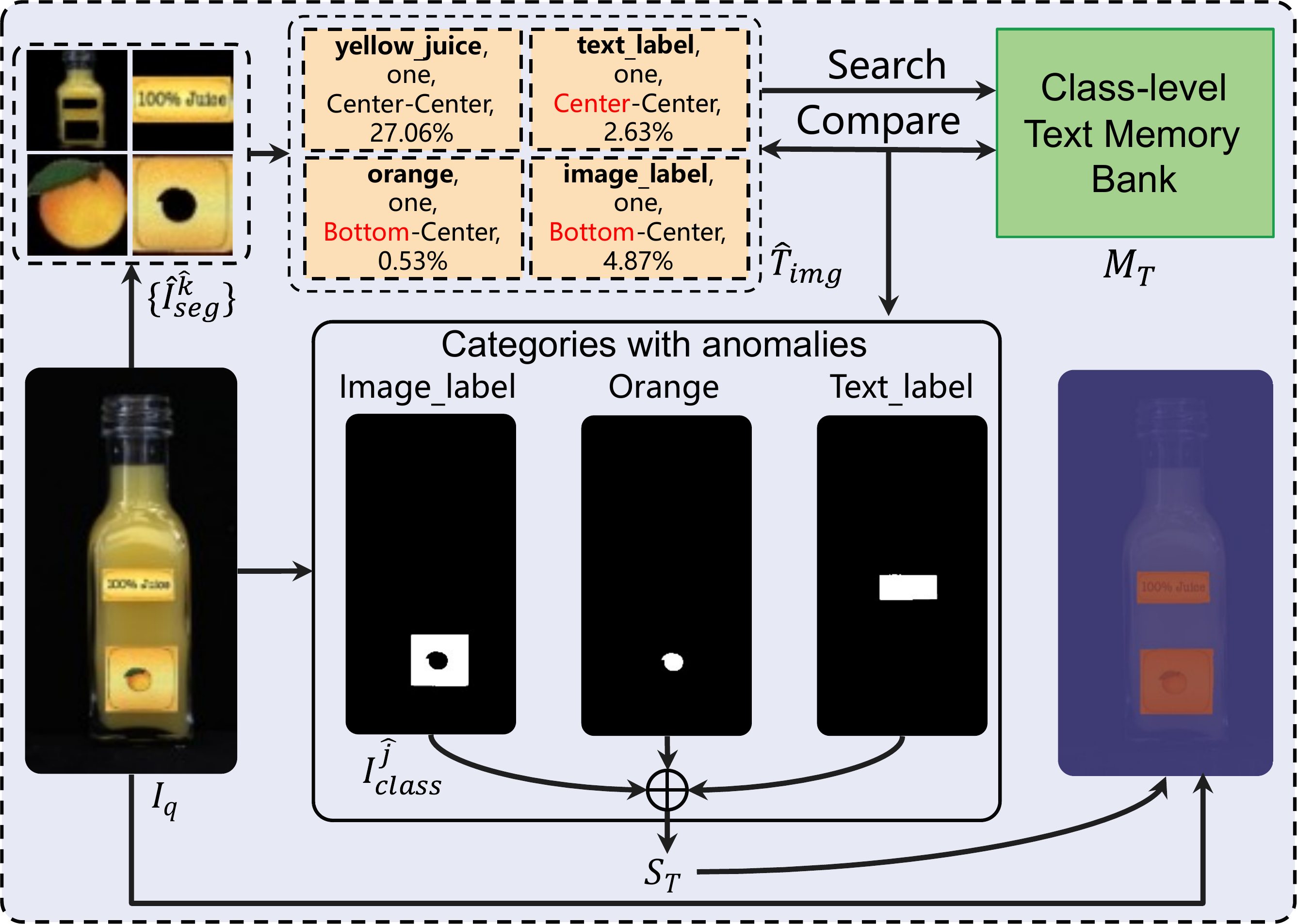} 
   \caption{
   The proposed pixel-level anomaly score calculation method for class-level text memory banks. A query image $I_q$ is first processed by the logic-aware text extractor to obtain $\hat{T}_{img}$. For each object category, the corresponding anomaly region $I_{class}^{\hat{j}}$ is identified by retrieving the most similar text description $T_{sim}$ of normal images. These regions $I_{class}^{\hat{j}}$ are merged to compute the class-level text anomaly score $S_T$.
   }
   \label{anomaly_score}
   \end{figure}
\begin{algorithm}[t]
   \SetKwData{Left}{left}\SetKwData{This}{this}\SetKwData{Up}{up}
   \SetKwFunction{Statistical}{Statistical}\SetKwFunction{return}{return}\SetKwFunction{similarity}{similarity}\SetKwFunction{Text}{Text}\SetKwFunction{Length}{Length}\SetKwFunction{Find}{Find}\SetKwFunction{diff}{diff}
   \SetKwInOut{Input}{Input}\SetKwInOut{Output}{Output}\SetKwFunction{max}{max}
   \SetKwFunction{Search}{Search}\SetKwFunction{MaxSize}{MaxSize}\SetKwFunction{MinSize}{MinSize}\SetKwFunction{Algorithm}{Algorithm}
   \Input{A query image ${I}_q$ in the testing set. The text memory bank $M_{T}$. $s_{min}^{\hat{j}}$ and $s_{max}^{\hat{j}}$ denote the maximum and minimum sizes of the object category $O_{class}^{\hat{j}}$ in ${I}_q$. The mask images $I_{c}$ ($c=O_{class}^t$) indicate all possible regions where the categories $O_{class}^t$ could occur.}
   \Output{Abnormal scores $S_{T}$ using text memory bank.}
   \BlankLine
   $\hat{T}_{img}, \{I_{class}^{\hat{j}}\}_{{\hat{j}}=1}^{\hat{n}_j} \leftarrow  \Algorithm 1(I_q)$\\
   $\{O_{class}^{\hat{j}},O_{num}^{\hat{j}},O_{pos}^{\hat{j}},O_{size}^{\hat{j}}\}_{{\hat{j}}=1}^{\hat{n}_j}\leftarrow \hat{T}_{img} $\\
   $T_{sim} \leftarrow$ \Search($T_{img}^q,M_T$)\\
   $\{O_{class}^t,O_{num}^t,O_{pos}^t,O_{size}^t\}_{t=1}^{n_t}\leftarrow T_{sim} $\\
   $S_T\leftarrow0$\\
   \For{${\hat{j}}\leftarrow 1$ \KwTo $n_j$}{
      \If{$O_{class}^{\hat{j}}$ \bf{not} \bf{in} $\{O_{class}^t\}_{t=1}^{n_t}$}{
      $S_T\leftarrow \max(S_T,I_{class}^{\hat{j}})$\Comment{categories}
      }
      \Else{
      \For{$t\leftarrow 1$ \KwTo $n_t$}{
         \If{$O_{class}^{\hat{j}} == O_{class}^t$}{
         \uIf{$O_{num}^{\hat{j}} \neq O_{num}^t$ \bf{or} $O_{pos}^{\hat{j}} \neq O_{pos}^t$}{
            $S_T\leftarrow \max(S_T,I_{class}^{\hat{j}})$\Comment{num, pos}
         }
         \uElseIf{$O_{size}^{\hat{j}} >s_{max}^{\hat{j}}$}{
         $\alpha =\left | O_{size}^{\hat{j}}-s_{max}^{\hat{j}} \right | /(s_{max}^{\hat{j}}-s_{min}^{\hat{j}})$ \\
         $S_T\leftarrow \max(S_T, \alpha \cdot I_{class}^{\hat{j}})$
         }
         \ElseIf{$O_{size}^{\hat{j}} < s_{min}^{\hat{j}}$}{
         $\alpha =\left | O_{size}^{\hat{j}}-s_{min}^{\hat{j}} \right | /(s_{max}^{\hat{j}}-s_{min}^{\hat{j}})$ \\
         $S_T\leftarrow \max(S_T, \alpha \cdot I_{class}^{\hat{j}})$\Comment{sizes}
         }
         }
      }}
   }
   \tcp{Detect missing categories.}
   \For{$t\leftarrow 1$ \KwTo $n_t$}{
   \If{$O_{class}^t$ \bf{not} \bf{in} $\{O_{class}^{\hat{j}}\}_{{\hat{j}}=1}^{\hat{n}_j}$}{
   $S_T\leftarrow \max(S_T, I_{c})$\Comment{$c=O_{class}^t,c \in C$}}
   }
   \caption{Class-level Text Anomaly Score.}\label{Algorithm2}
\end{algorithm}

\begin{figure*}[b]
   \centering
   \includegraphics[width=0.92\textwidth]{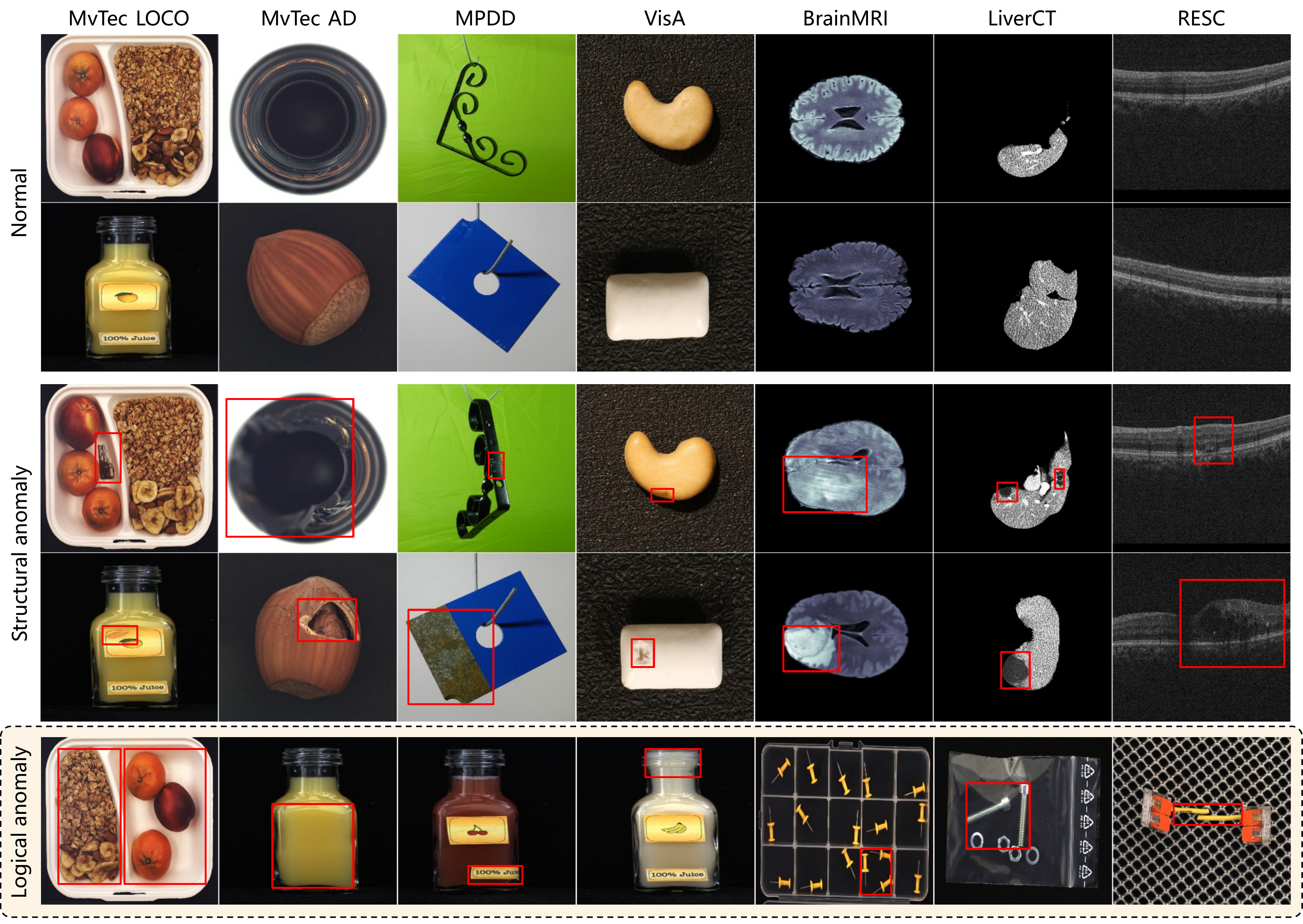}
   \caption{Representative examples from industrial \cite{bergmann2022beyond,bergmann2019mvtec,9631567,zou2022spot} and medical datasets \cite{baid2021rsna,landman2015miccai,HU2019216} for both structural and logical anomalies with anomalous regions highlighted in red.}
   \label{dataset_sample}
   \end{figure*}

As shown in Fig. \ref{anomaly_score} and Algorithm~\ref{Algorithm2}, we first search the class-level text memory bank $M_T$ for the text $T_{sim}$ that is most similar to the text description $\hat{T}_{img}$ of the query image $I_q$ \cite{wikipedia_gestalt}. We then compute anomaly scores by comparing each object categories in $T_{sim}=\{O_{class}^t,O_{num}^t,O_{pos}^t,O_{size}^t\}_{t=1}^{n_t}$ with those in $\hat{T}_{img}:=\{O_{class}^{\hat{j}},O_{num}^{\hat{j}},O_{pos}^{\hat{j}},O_{size}^{\hat{j}}\}_{{\hat{j}}=1}^{\hat{n}_j}$, where $n_t$ and $\hat{n}_j$ denote the total number of object categories in $T_{sim}$ and $\hat{T}_{img}$, respectively. For anomalies in object category $O_{class}^{\hat{j}}$, count $O_{num}^{\hat{j}}$, and location $O_{pos}^{\hat{j}}$, we use the merged mask $I_{class}^{\hat{j}}$ of all objects belonging to category $O_{class}^{\hat{j}}$ as the pixel-level anomaly score, i.e., $S_T= \mathrm{max}(S_T,I_{class}^{\hat{j}})$, where $\mathrm{max}(\cdot)$ denotes a pixel-wise maximum operator. $S_T$ is initialized as a zero image with the same width and height as $I_q$. For abnormal in object sizes $O_{size}^{\hat{j}}$, we calculate a coefficient $\alpha$ that quantifies the deviation of $O_{size}^{\hat{j}}$ from the normal size range $\left [ s_{min}^{\hat{j}},s_{max}^{\hat{j}} \right ] $, i.e., $S_T= \mathrm{max}(S_T,\alpha \cdot I_{class}^{\hat{j}})$.
$s_{min}^{\hat{j}}$ and $s_{max}^{\hat{j}}$ denote the maximum and minimum sizes of the object category $O_{class}^{\hat{j}}$ in the memory bank $M_T$.
\begin{equation}
s_{max}^{\hat{j}}=\mathop{\max}_{T_{img}^i\in M_T}\left \{ O_{size}^j| O_{class}^j=O_{class}^{\hat{j}}\right \},
\end{equation}
\begin{equation}
s_{min}^{\hat{j}}=\mathop{\min}_{T_{img}^i\in M_T}\left \{ O_{size}^j| O_{class}^j=O_{class}^{\hat{j}}\right \},
\end{equation}
where $T_{img}^i=\{O_{class}^j,O_{num}^j,O_{pos}^j,O_{size}^j\}_{j=1}^{n_j}$.
To detect the missing object categories $O_{class}^t$ in the query image $I_q$, we record all possible locations where the categories $O_{class}^t$ could appear, denoted as $I_{c}$ in Algorithm~\ref{Algorithm1}.
$I_{c}$ are used to compute the pixel-level anomaly score corresponding to the missing object categories, i.e., $S_T= \mathrm{max}(S_T, I_{c})$.

\subsubsection{Object-level Image Anomaly Score}
We use SAM \cite{kirillov2023segment} to segment the query image $I_q$ into different objects $\hat{I}_{seg}^{\hat{s}}$ and use CLIP \cite{radford2021learning} to extract the object-level image features $\hat{F}_O=\{\hat{F}_{seg}^{\hat{s}}\}_{\hat{s}=1}^{\hat{n}_s}$, $\hat{F}_{seg}^{\hat{s}}=E_{clip}(\hat{I}_{seg}^{\hat{s}})$, where $\hat{n}_s$ is the number of segmented objects in the query image $I_q$ by SAM \cite{kirillov2023segment}. The object-level image anomaly score $S_O$ can be calculated by
\begin{equation}  
S_O= {\textstyle \sum_{\hat{s}=1}^{\hat{n}_s}} (1-  \mathop{\max}_{F_{seg}^s \in F_O^i, F_O^i \in M_O}(\frac{F_{seg}^s\cdot {\hat{F}_{seg}^{\hat{s}}}{}^\top}{\Vert  F_{seg}^s \Vert \Vert \hat{F}_{seg}^{\hat{s}} \Vert }))\cdot \hat{I}_{seg}^{\hat{s}}.\label{eq:seg_pix_score}
\end{equation}

\subsubsection{Patch-level Anomaly Score}
We employ CLIP \cite{radford2021learning} and DINOv2 \cite{oquab2023dinov2} to extract patch-level image features $\hat{F}_P$ from the query image $I_q$.
The patch-level anomaly score $S_P$ can be calculated by 
\begin{equation}  
  {S}_{P} = \mathop{\max}_{F_P^i \in M_P}(1-\text{upsample}(\frac{F_{P}^i\cdot \hat{F}_{P}^\top}{\Vert F_{P}^i\Vert \Vert \hat{F}_{P}\Vert})),
\label{eq:patch_pixel_score}
\end{equation}
where $\mathrm{upsample}(\cdot)$ denotes upsampling the patch feature map to the resolution of the query image $I_q$.

\subsubsection{Anomaly Score Fusion}
Considering that $M_{P}$ and $M_{O}$ focus on structural anomaly detection, while $M_{T}$ are designed to detect logical anomalies, we perform a weighted fusion of $S_T, S_I, S_P$ to derive the pixel-level anomaly score map, 
\begin{equation}  
S_{pixel}=\lambda_1 S_{T}+\lambda_2 S_{I}+\lambda_3 S_{P}.    
\label{eq:all_score}
\end{equation}
The fusion weights are empirically set as $\lambda_1 = 0.05$, $\lambda_2 = 0.3$, and $\lambda_3 = 0.65$.
Finally, we take the maximum value of ${S}_{pixel}$ as the image-level anomaly score ${S}_{image}$.

\begin{figure*}[b]
\centering
\includegraphics[width=0.95\textwidth]{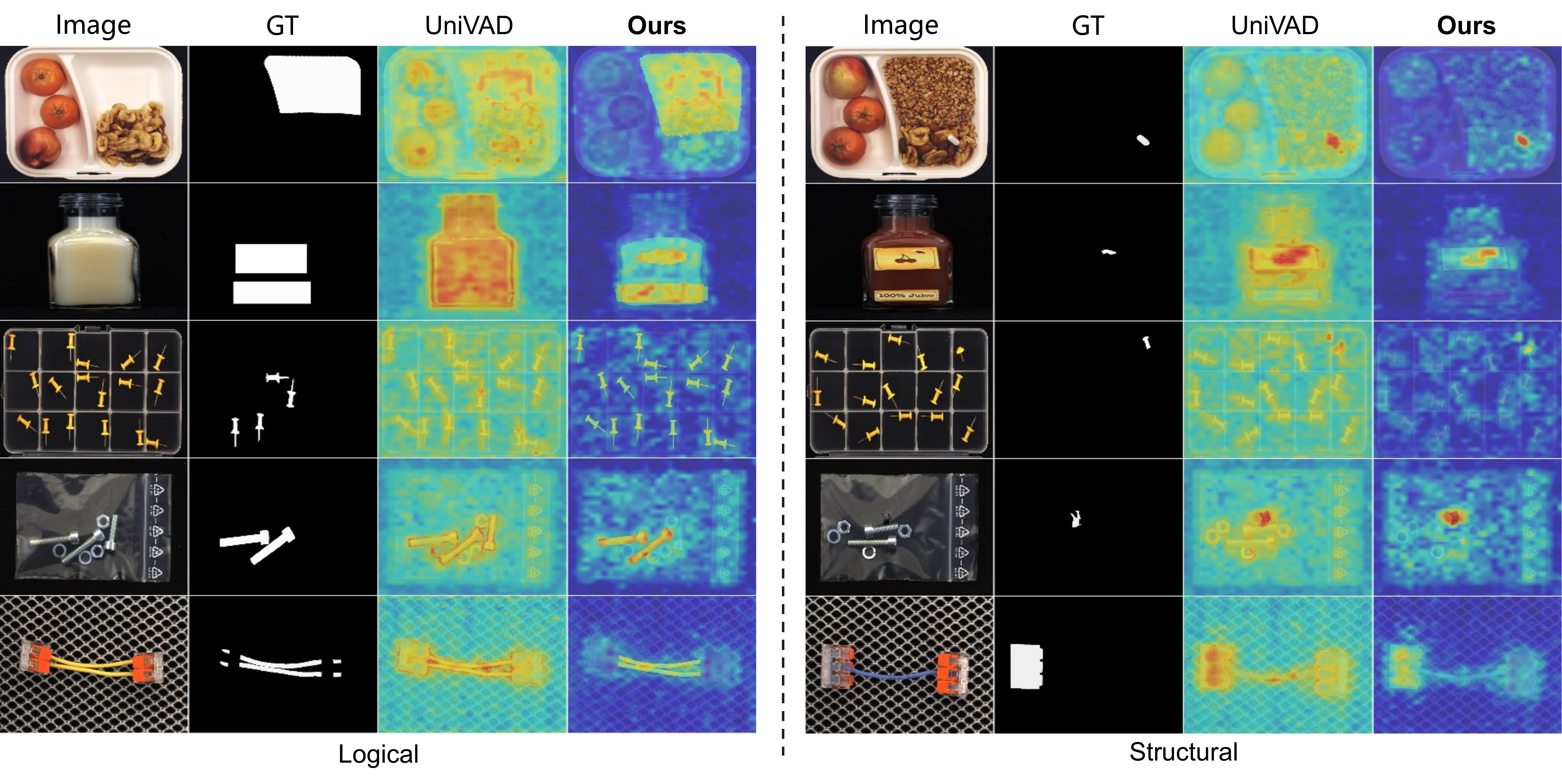} 
\caption{Visual comparison results on the MVTec LOCO \cite{bergmann2022beyond} dataset.}
\label{mvtecloco_location}
\end{figure*}

\begin{table*}[b]
   \caption{Image-level AUROC comparison results on MVTec LOCO \cite{bergmann2022beyond} dataset for unified logical and structural anomaly detection.
   The results marked ``$\ast$'', ``$\P$'' and ``$\ddagger$'' are reported by~\cite{zhang2025towards},~\cite{kim2024few} and~\cite{zhao2024logical}, respectively. ``Individual CLIP" indicates CLIP trained individually on each dataset.}
   \footnotesize
   \begin{center}
   \resizebox{\textwidth}{!}{
   \begin{tabular}{c|c|c|c|c|c|c|c|c|c|c||c|c}
   \hline
   \multirow{2}*{Category} 
   & Dream $\P$ & GCAD$\ast$  & PatchCore $\ast$ 
   & OmniAL  $\ddagger$ & GeneralAD$\ast$ 
   & LogicAL & PBAS  & UniVAD  & LogSAD  
   & \bf{Ours} & \multicolumn{2}{c}{Individual CLIP}\\ 
   & \cite{zavrtanik2021draem} CVPR21 & \cite{bergmann2022beyond} IJCV21 & \cite{roth2022towards} CVPR22 & \cite{zhao2023omnial} CVPR23 & \cite{strater2024generalad} ECCV24 & \cite{zhao2024logical}  CVPR24 &\cite{10716437} TCSVT25 & \cite{gu2025univad} CVPR25 & \cite{zhang2025towards} CVPR25 & TMUAD-U & TMUAD-I & PBAS+$M_T$
   \\ \hline
   {Breakfast Box} & 80.3 & 83.9 & 77.1 & 75.9 & - & 85.4 & \underline{88.4} & \underline{88.4} & \underline{95.7} & \textbf{99.5} & 99.5 & 94.5\\
   {Juice Bottle}  & 94.3 & \underline{99.4} & 94.6 & \textbf{99.5} & - & 98.5 & 89.1 & 90.9 & 95.2 & 94.9 & 94.9 & 93.1\\
   {Pushpins}      & 68.6 & 86.2 & 74.1 & 79.6 & - & \underline{87.4} & 76.5 & 74.9 & 83.6 & \textbf{93.1} & 93.1 & 91.3\\
   {Screw Bag}     & 70.6 & 63.2 & 73.3 & 83.1 & - & 82.0 & 73.2 & 66.5 & \underline{83.2} & \textbf{94.9} & 96.9 & 94.3\\
   {Splicing Connectors} & 85.4 & 83.9 & 86.0 & 88.1 & - & 89.0 & 80.0 & 88.8 & \textbf{93.5} & \underline{91.2} & 91.8 & 86.8\\ \hline
   $\Delta$AUROC & 25.7 & 36.2 & 21.3 & 23.6 & - & 16.5 & 15.9 & 24.4 & \underline{12.5} & \textbf{8.3} & 7.7 & 7.7\\
   {\textbf{Average}} & 79.8 & 83.3 & 81.0 & 85.0 & 84.9 & 88.5 & 81.4 & 81.9 & \underline{90.2} & \textbf{94.6} & 95.2 & 92.0\\ \hline
   \end{tabular}
   }
   \end{center}
   \label{mvtec_loco}
   \end{table*}

\section{Experiments}
\subsection{Dataset setting and Evaluation Metrics}
We evaluated our method on two anomaly detection tasks, including logical anomaly detection and structural anomaly detection on the widely used 7 anomaly detection datasets in industrial and medical domains. 
The area under the receiver operating characteristic curve (AUROC) is employed as the primary evaluation metric.
For all datasets, the training set contains only normal samples, while the test set contains both normal and abnormal samples. Representative examples are shown in Fig. \ref{dataset_sample}, where abnormal regions are marked with red bounding boxes.

For {\bf industrial logical anomaly detection and structural anomaly detection}, we use the MVTec LOCO dataset \cite{bergmann2022beyond}, containing 1772 normal samples for training and 1568 samples (575 normal and 993 anomalous sample) for testing.

\begin{table*}[!t]
   \caption{Image-level AUROC comparison results on industrial and medical anomaly detection datasets.
   The results marked ``$\ast$'' and ``$\P$'' are reported by~\cite{ma2025aa} and~\cite{gu2025univad}, respectively. ``Individual CLIP" indicates CLIP trained individually on each dataset.}
   \footnotesize
   \begin{center}
   \resizebox{\textwidth}{!}{
   \begin{tabular}{c|c|c|c|c|c|c|c|c|c||c|c}
   \hline
   \multirow{2}*{Dataset} & WinCLIP $\ast$ & VAND $\ast$ & AnomalyGPT $\P$ & AnomalyCLIP $\ast$ & MVFA-AD $\ast$ & AdaCLIP $\ast$ & AA-CLIP  & UniVAD  & \bf{Ours} &\multicolumn{2}{c}{Individual CLIP}\\ 
   & \cite{jeong2023winclip} CVPR23 & \cite{chen2023april} CVPRW23 & \cite{gu2024anomalygpt} AAAI24 & \cite{zhou2023anomalyclip} ICLR24 & \cite{huang2024adapting} CVPR24 & \cite{cao2024adaclip} ECCV24 & \cite{ma2025aa} CVPR25 & \cite{gu2025univad} CVPR25&TMUAD-U&TMUAD-I& PBAS+$M_T$ / PBAS\\ \hline
   MVTec AD \cite{bergmann2019mvtec} & 91.8 & 86.1 & \underline{94.1} & 90.9 & 86.6 & 90.0 & 90.5 & \textbf{99.0} & \textbf{99.0} & 99.1 & 99.5/99.5\\
   VisA \cite{zou2022spot} & 78.0 & 66.4 & 87.4 & 82.1 & 76.5 & 84.3 & 84.6 & \textbf{96.2} & \underline{95.9} & 96.0 & 97.1/97.1\\
   MPDD \cite{9631567} & 63.6 & 73.0 & - & 73.7 & 70.9 & 72.1 & 75.1 & \underline{97.2} & \textbf{97.5} & 98.3 & 97.9/97.9\\
   BrainMRI \cite{baid2021rsna} & 66.5 & 58.8 & 73.2 & 83.3 & 70.9 & {80.2} & 80.2 & \underline{84.2} & \textbf{89.2}& - & -\\
   RESC \cite{HU2019216} & 42.5 & 65.6 & 82.4 & 75.7 & 77.3 & 82.7  & 82.7& \underline{89.4} & \textbf{89.6}& - & -\\
   LiverCT \cite{landman2015miccai} & 64.2 & 54.7 & 60.3 & 61.6 & 63.0 & 64.2 & 69.7 & \underline{70.7} & \textbf{72.1}& - & - \\ \hline
   \end{tabular}
   }
   \end{center}
   \label{mvtec_ad}
   \end{table*}

   \begin{table*}[!t]
   \caption{Pixel-level AUROC comparison results on industrial and medical anomaly detection datasets.
   }
   \footnotesize
   \begin{center}
   \resizebox{\textwidth}{!}{
   \begin{tabular}{c|c|c|c|c|c|c|c|c|c||c|c}
   \hline
   \multirow{2}*{Dataset} & WinCLIP $\ast$ & VAND $\ast$ & AnomalyGPT $\P$ & AnomalyCLIP $\ast$ & MVFA-AD $\ast$ & AdaCLIP $\ast$ & AA-CLIP  & UniVAD  & \bf{Ours} &\multicolumn{2}{c}{Individual CLIP}\\ 
   & \cite{jeong2023winclip} CVPR23 & \cite{chen2023april} CVPRW23 & \cite{gu2024anomalygpt} AAAI24 & \cite{zhou2023anomalyclip} ICLR24 & \cite{huang2024adapting} CVPR24 & \cite{cao2024adaclip} ECCV24 & \cite{ma2025aa} CVPR25 & \cite{gu2025univad} CVPR25&TMUAD-U&TMUAD-I& PBAS+$M_T$ / PBAS\\ \hline
   MVTec AD \cite{bergmann2019mvtec} & 85.1 & 87.6 & 95.3 & 91.1 & 84.9 & 89.9 & 91.9 & \underline{96.3} & \textbf{96.9} & 96.9 & 98.3/98.3\\
   VisA \cite{zou2022spot} & 79.6 & 94.2 & 96.2 & 95.4 & 93.4 & 95.5& 95.5 & \underline{98.5} & \textbf{98.8} & 98.8 & 94.9/94.9\\
   MPDD \cite{9631567} & 95.2 & 94.9 & - & 96.2 & 94.5 & 96.6 & 96.7 & \underline{97.7} & \textbf{98.5} & 98.6 & 97.8/97.8\\
   BrainMRI \cite{baid2021rsna} & 86.0 & 94.5 & 96.0 & 96.2 & 95.6 & 93.9 & 95.5 & \underline{96.7} & \textbf{97.6} &- & -\\
   LiverCT \cite{landman2015miccai} & 96.2 & 95.6 & 95.8 & 93.9 & 96.8 & 94.5 & \textbf{97.8} & 97.2 & \underline{97.6} &- & - \\ \hline
   \end{tabular}
   }
   \end{center}
   \label{mvtec_ad_pix}
   \end{table*}
   
\begin{figure*}[b]
   \centering
   \includegraphics[width=0.98\textwidth]{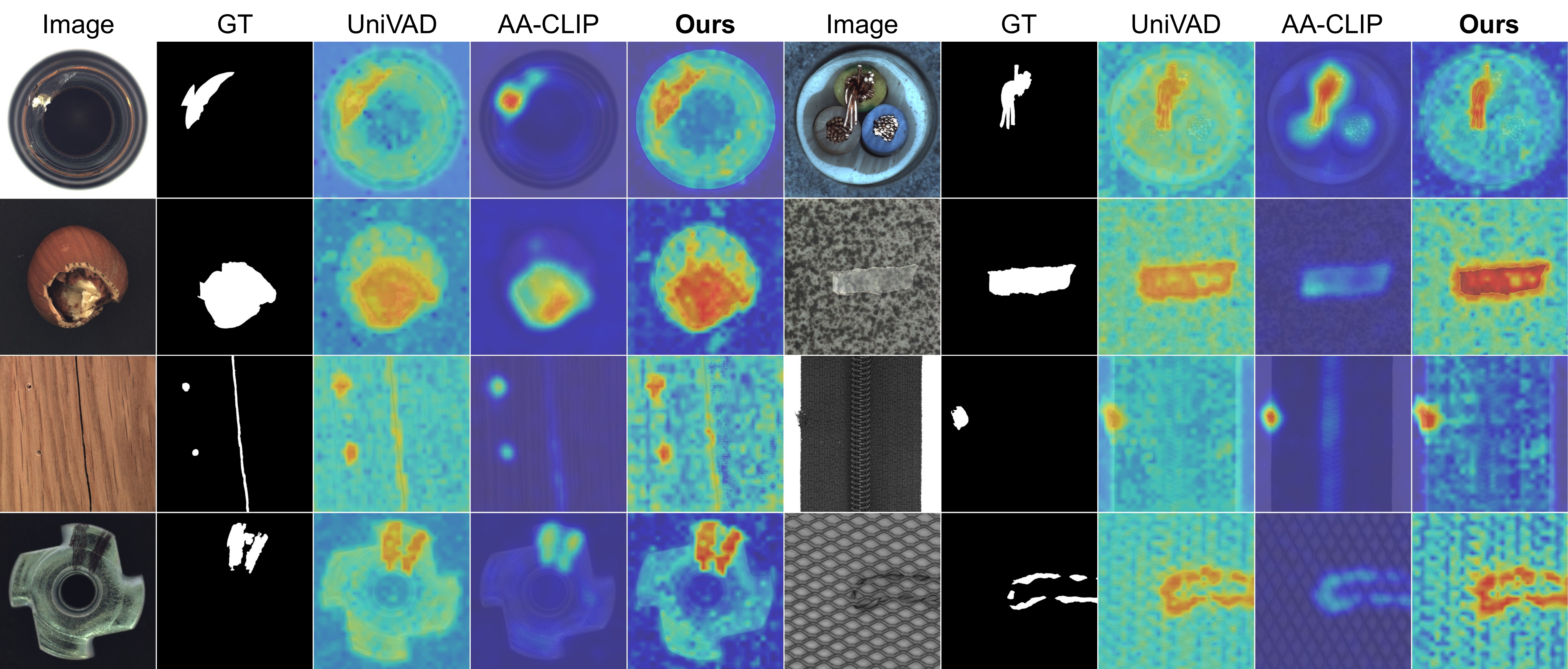} 
   \caption{Visual comparison results on the MVTec AD \cite{bergmann2019mvtec} dataset.}
   \label{mvtec_ad_location}
\end{figure*}
      
\begin{figure*}[!t]
\centering
\includegraphics[width=0.98\textwidth]{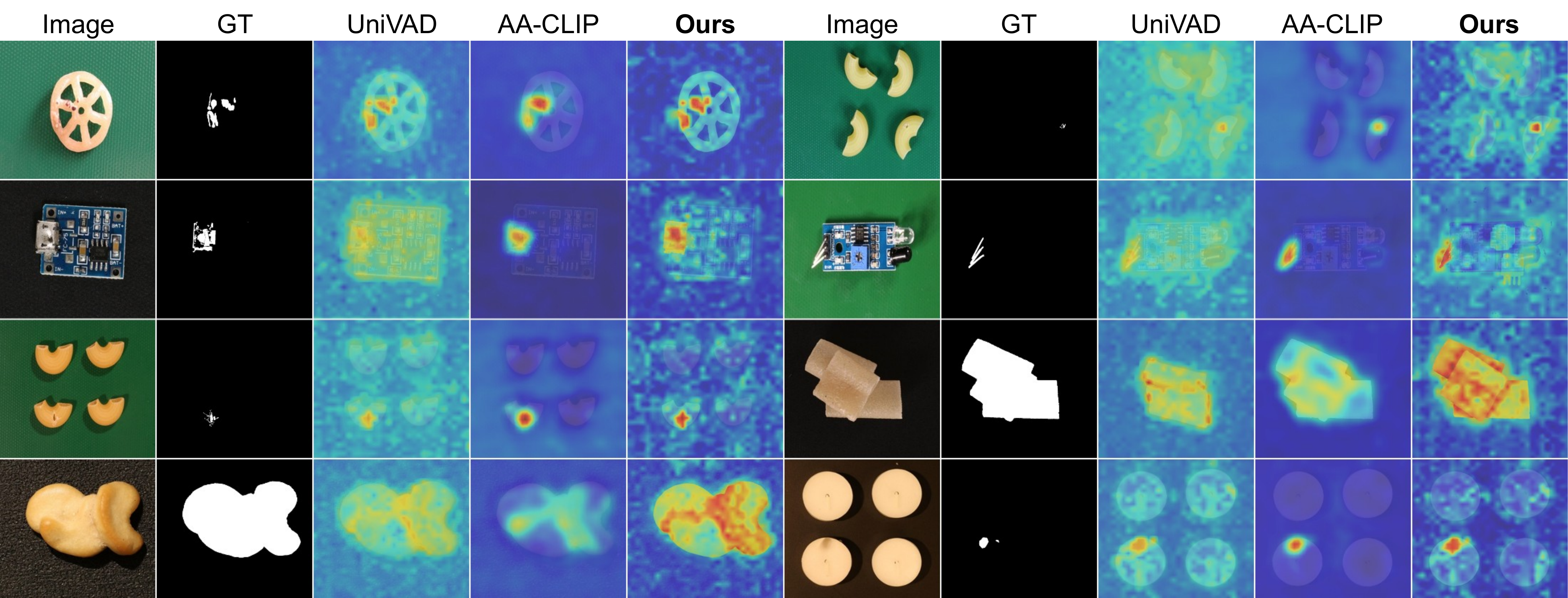} 
\caption{Visual comparison results on the VisA \cite{zou2022spot} dataset.}
\label{VisA_ad_location}
\end{figure*}

For the {\bf industrial structural anomaly detection}, we employ three widely used datasets: MVTec AD \cite{bergmann2019mvtec}, MPDD \cite{9631567}, and VisA \cite{zou2022spot}. These datasets primarily consist of structural anomalies such as scratches, dents, contaminations, and various structural changes. MVTec AD  simulates real industrial production scenarios, containing 3629
normal samples for training and 1725 samples (467 normal and 1258 anomalous sample) for testing across 15 categories. MPDD focuses on metal part defect detection, with a training set of 888 normal samples and a test set comprising 176 normal and 281 anomalous samples. VisA introduces objects with complex structures and variations in pose or position, comprising 9621 normal samples for training and 1200 anomalous samples for test.

To evaluate the generalization ability of the proposed framework, we conducted experiments on three {\bf medical anomaly detection} datasets: BrainMRI \cite{baid2021rsna}, LiverCT \cite{landman2015miccai}, and RESC \cite{HU2019216}. The BrainMRI dataset is a human brain tumor segmentation benchmark commonly used in medical image analysis. It contains a training set with 7500 normal samples and a test set with 640 normal and 3,075 abnormal samples. The LiverCT dataset consists of abdominal CT scans. Its training set includes 1542 normal samples, while the test set comprises 833 normal and 660 abnormal samples. The RESC dataset targets ophthalmic disease detection. It provides 4,297 normal training samples and a test set with 1,041 normal and 764 abnormal samples.

\subsection{Implementation Details}
Before constructing the three memory banks, we employ Grounded SAM \cite{ren2024grounded} to detect and segment objects in normal images. Grounded SAM integrates Grounding DINO \cite{liu2024grounding} for object category detection and SAM \cite{kirillov2023segment} for object segmentation. The segmented objects from seven datasets are then organized by category, and both the categorized segmented images (from SAM \cite{kirillov2023segment}) and their corresponding text labels (from Grounding DINO \cite{liu2024grounding}) are used to fine-tune CLIP \cite{radford2021learning} (ViT-L/14). In this fine-tuning process,  200 images per category are used to enhance the classification accuracy of our logic-aware text extractor.

Except for the CLIP model in the text extractor, which uses our fine-tuned weights, all other models use publicly available pre-trained weights, including VLM \cite{wang2024qwen2} in Fig. \ref{Visual_Describer}, as well as CLIP \cite{radford2021learning}, DINOv2 \cite{oquab2023dinov2}, Grounding SAM \cite{ren2024grounded}, and SAM \cite{kirillov2023segment} in Fig. \ref{main_framework_diagram}.
For all datasets, input images for all models are uniformly resized to 448×448 pixels. The number of clusters in the K-means algorithm is set to 100 for each image category to build patch-level image memory bank $M_P$ and is set to 1000 for each object category to build object-level image memory bank $M_O$.

\subsection{Comparison with State-of-the-art Methods}
For fair comparisons, we retrained UniVAD \cite{gu2025univad} on full data and PBAS \cite{10716437} using publicly available code on seven datasets containing industrial anomalies \cite{bergmann2022beyond,bergmann2019mvtec,9631567,zou2022spot} and medical anomalies \cite{baid2021rsna,landman2015miccai,HU2019216}, while the results of other methods are provided from the original papers as much as possible. Except for ``TMUAD-I'' and ``PBAS+$M_T$'' in Table \ref{mvtec_loco}, \ref{mvtec_ad}, and \ref{mvtec_ad_pix}, which use CLIP trained individually on each dataset, we use CLIP with the same weight in the text extractor is applied across all datasets to achieve unified logical and structural anomaly detection.

For unified logical anomaly detection and structural
anomaly detection, we compare our TMUAD with the following methods on the MVTec LOCO \cite{bergmann2022beyond} dataset:  PatchCore \cite{roth2022towards}, Dream \cite{zavrtanik2021draem}, PBAS \cite{10716437}, UniVAD \cite{gu2025univad}, GCAD \cite{bergmann2022beyond}, GeneralAD \cite{strater2024generalad}, OmniAL \cite{zhao2023omnial}, LogicAL \cite{zhao2024logical}, LogSAD \cite{zhang2025towards} in full-data evaluation results. 
Table \ref{mvtec_loco} shows that our method substantially outperforms state-of-the-art methods in terms of image-level AUROC (94.6). Unlike existing methods, which exhibit large performance fluctuations across categories, our method maintains balanced performance across different image categories, demonstrating greater robustness in anomaly detection.
Table \ref{mvtec_loco_LA_only} further confirms that our method achieves the SOTA image-level AUROC for both logical and structural anomaly detection.
Fig. \ref{mvtecloco_location} illustrates that TMUAD accurately localizes anomaly regions with minimal noise, consistently across both logical and structural anomalies.

\begin{table}[!h]
\caption{Image-level AUROC comparison results on the MVTec LOCO\cite{bergmann2022beyond} dataset. “Average” represents the image-level average AUROC for unified logical and structural anomaly detection.}
\centering
\small
\begin{tabular}{c|c|c|c}
\hline
Methods & Logical & Structural & Average \\ \hline
SimpleNet \cite{10203746} & 71.5 & 83.7 & 77.6 \\
PBAS \cite{10716437} & 76.6 & 88.5 & 82.6 \\
UniVAD \cite{gu2025univad} & 73.8 & 92.9 & 83.3 \\ 
LogSAD \cite{zhang2025towards} & \underline{89.3} & \underline{93.1} & \underline{91.2} \\ \hline
\textbf{Ours} & \textbf{91.8} &  \textbf{98.5} & \textbf{95.2} \\ \hline
\end{tabular}
\label{mvtec_loco_LA_only}
\end{table}

\begin{figure}[t]
\centering
\includegraphics[width=\columnwidth]{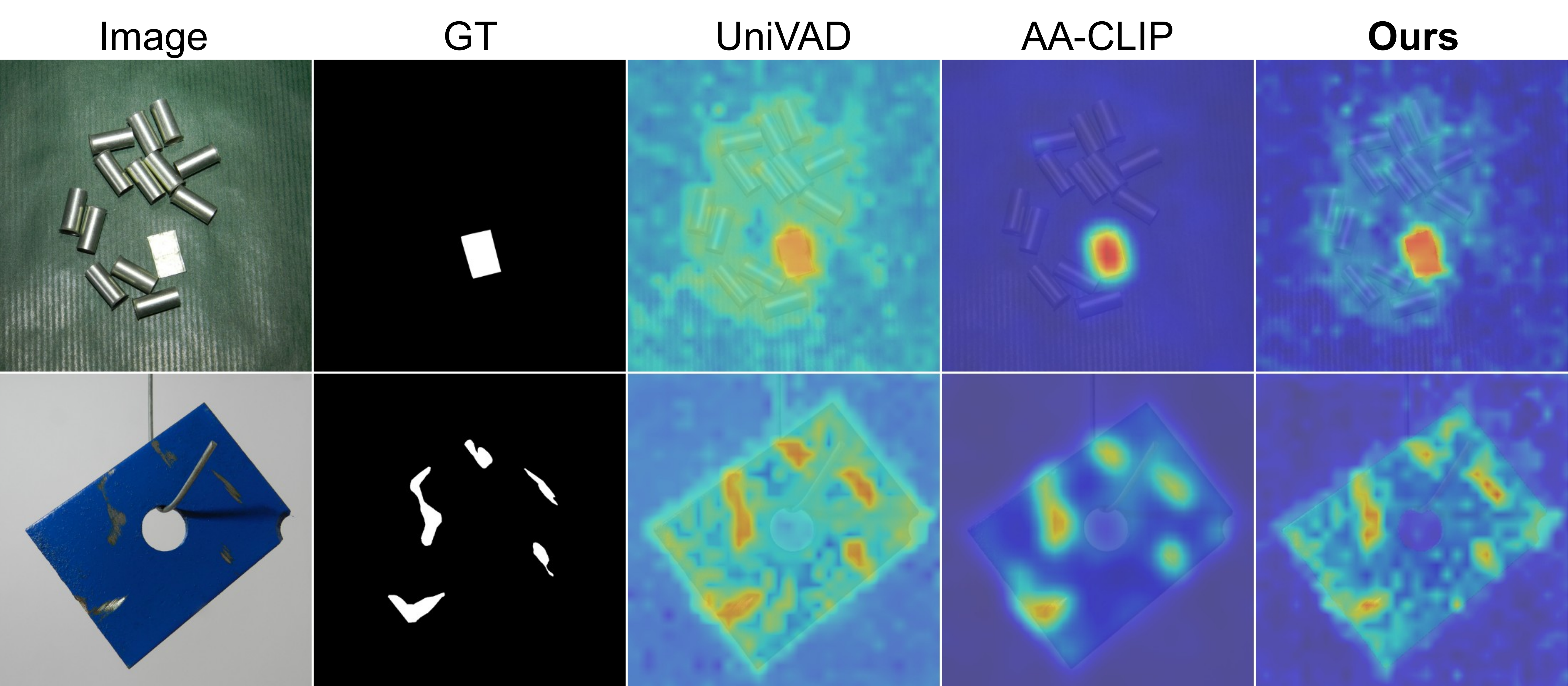} 
\caption{Visual comparison results on  the MPDD \cite{9631567} dataset.}
\label{mpdd_result}
\end{figure}

\begin{figure}[t]
\centering
\includegraphics[width=\columnwidth]{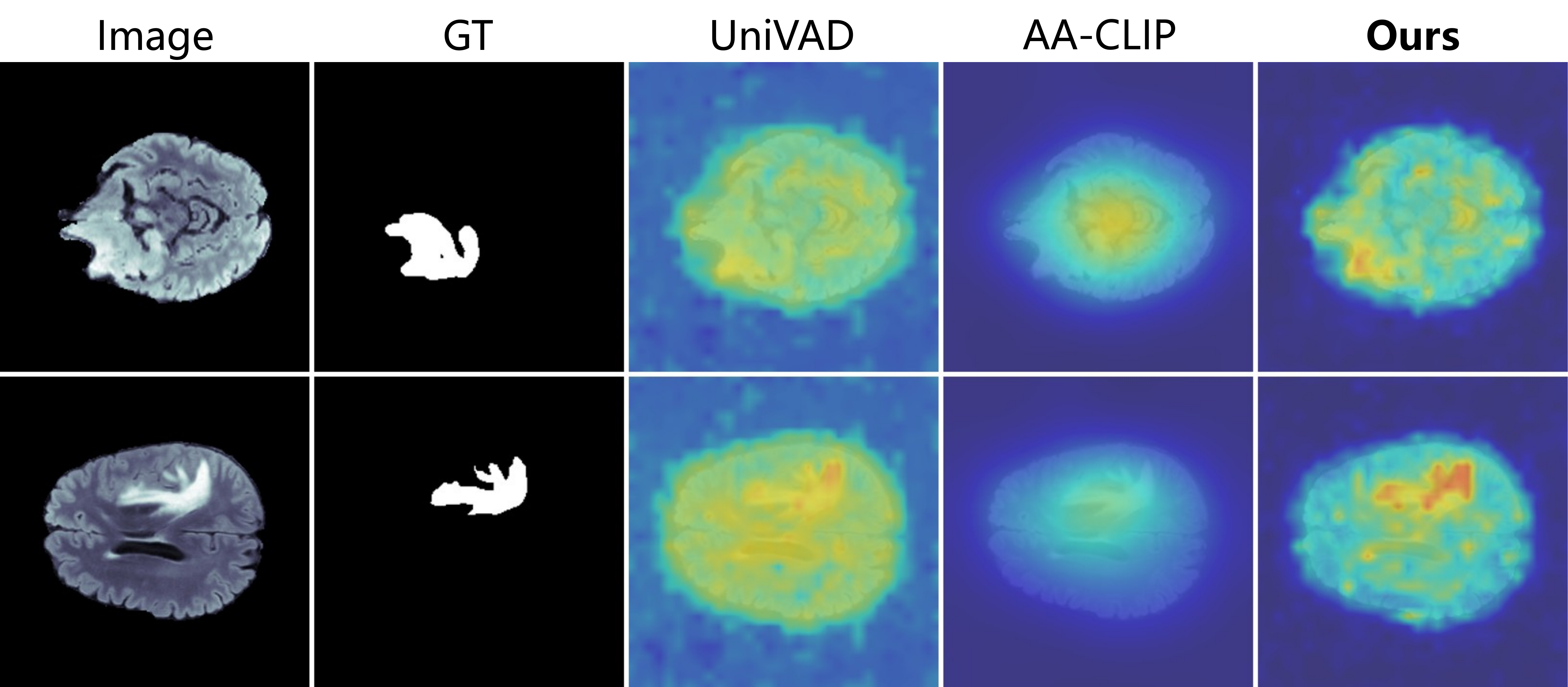} 
\caption{Visual comparison results on the BrainMRI \cite{baid2021rsna} dataset compared with other methods.}
\label{BrainMRI_result}
\end{figure}

For structural anomaly detection, we compare our TMUAD with the following method on 3 industrial structural datasets (MVTec AD \cite{bergmann2019mvtec} in Fig. \ref{mvtec_ad_location}, VisA  \cite{zou2022spot} in Fig. \ref{VisA_ad_location}, and MPDD   \cite{9631567} in Fig. \ref{mpdd_result}) and 3 medical datasets (BrainMRI \cite{baid2021rsna} in Fig. \ref{BrainMRI_result}, RESC \cite{HU2019216}, and LiverCT \cite{landman2015miccai}): WinCLIP \cite{jeong2023winclip}, VAND \cite{chen2023april}, AnomalyGPT \cite{gu2024anomalygpt}, AnomalyCLIP \cite{zhou2023anomalyclip}, MVFA-AD \cite{huang2024adapting}, AdaCLIP \cite{cao2024adaclip}, AA-CLIP \cite{ma2025aa}, and UniVAD \cite{gu2025univad}.
We retrained UniVAD \cite{gu2025univad} on full data for fair comparisons.
Table \ref{mvtec_ad} and Table \ref{mvtec_ad_pix} show the image-level and pixel-level AUROC metrics for different datasets, respectively. 

Table \ref{mvtec_loco}, \ref{mvtec_ad}, and \ref{mvtec_ad_pix} demonstrate that state-of-the-art anomaly detection performance can be achieved using unified weights. Furthermore, by applying dataset-specific weight of CLIP in the text extractor, we can obtain state-of-the-art results within a unified framework. For example, ``PBAS+$M_T$" preserves its strong structural anomaly detection capability while simultaneously achieving competitive performance on logical anomaly detection.

\begin{table}[h]
   \caption{Ablation studies of three memory bank on the MVTec LOCO \cite{bergmann2022beyond} dataset.}
   \centering
   \small
   \resizebox{\columnwidth}{!}{
   \begin{tabular}{ccc|c|c|c|c|c|c}
   \hline
   \toprule
   \multicolumn{3}{c|}{Memory banks}& \multicolumn{2}{c|}{Structural} & \multicolumn{2}{c|}{Logical}& \multicolumn{2}{c}{Average} \\ 
   \hline
   $M_{T}$ & $M_{O}$ & $M_{P}$ & Image & Pixel & Image & Pixel & Image & Pixel\\ \hline
   \checkmark &  &  & 63.4 & 64.0 & \underline{88.4} & \underline{77.6}&75.9& 70.8\\
    & \checkmark &  & \underline{96.4} & 62.0 & 58.1 & 54.1 & 77.3 &58.1\\
   &  & \checkmark & 91.1 & 91.4 & 69.7 & 73.0& 80.4 & 82.2\\
   & \checkmark & \checkmark & \textbf{98.5} & \underline{94.2} & 70.4 & 76.1&\underline{84.5}&\underline{85.2} \\
   \checkmark & \checkmark & \checkmark & \textbf{98.5} & \textbf{94.6} & \textbf{91.8} & \textbf{84.8} &\textbf{95.2}& \textbf{88.3}\\ 
   \hline
   \end{tabular}
   }
   \label{Ablation}
   \end{table}

   \begin{table*}[t]
      \caption{Ablation studies of the class-level text memory bank $M_T$ on the MVTec LOCO \cite{bergmann2022beyond} dataset in image-level AUROC.}
      \centering
      \small
      \resizebox{\textwidth}{!}{
      \begin{tabular}{c|c|c|c|c|c|c}
      \hline
      \multicolumn{3}{c|}{Memory bank construction}&\multicolumn{3}{c|}{Anomaly score calculation} & AUROC \\
      \hline
      Grounded SAM \cite{ren2024grounded} & VLMs \cite{wang2024qwen2} $\to T_{img}^j$ & VLMs $\to O_{pos}^j\to T_{img}^j$ & CLIP text encoder\cite{radford2021learning} & T5 encoder\cite{raffel2020exploring}  & Class-level Text Matching &  only $M_T$ \\ \hline
      &\checkmark &  & \checkmark & & & 54.0\\
      &\checkmark &  &  &\checkmark & & 55.9\\
      \checkmark &\checkmark &  &  &\checkmark & & 57.5\\
      \checkmark & & \checkmark  &  &\checkmark & & \underline{81.7}\\
      \checkmark & & \checkmark  &  & &\checkmark & \textbf{88.4}\\
      \hline
      \end{tabular}
      }
      \label{Text_Ablation}
      \end{table*}

\subsection{Ablation Studies}
\begin{figure}[!t]
   \centering
   \includegraphics[width=0.98\columnwidth]{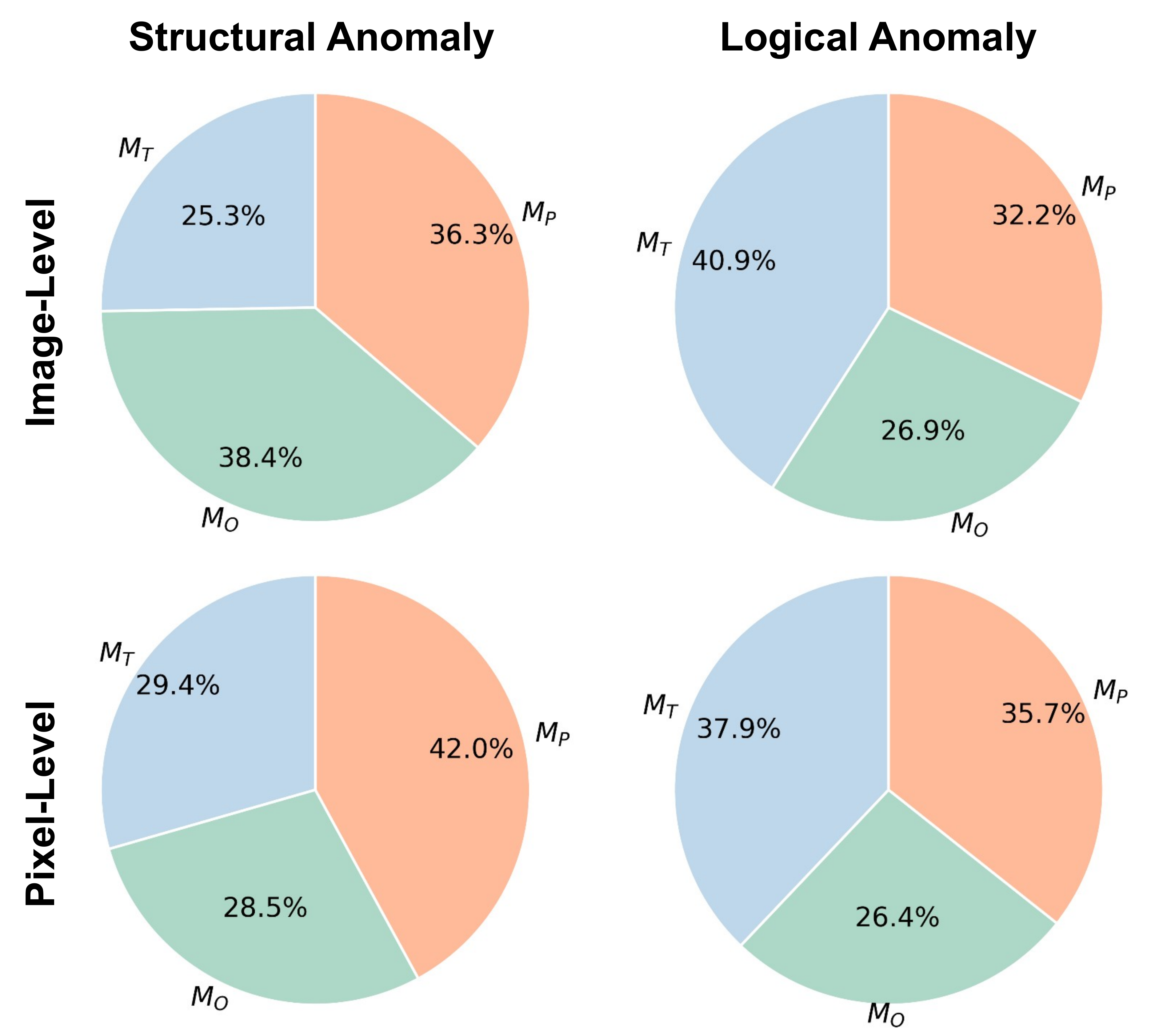} 
   \caption{The relative contributions of three distinct memory banks to the performance of structural and logical anomaly detection.}
   \label{Ablation_pie_charts_pixel}
   \end{figure}

We conducted ablation studies of our memory banks on the MVTec LOCO \cite{bergmann2022beyond} dataset, as shown in Table \ref{Ablation} and Fig. \ref{Ablation_pie_charts_pixel}. 
The image-level and pixel-level AUROC shown in Table~\ref{Ablation} are calculated using $M_T,M_O,M_P$.
We make three key observations: 
1) $M_O$ is most effective for image-level AUROC in detecting structural abnormalities;
2) $M_P$ achieves the highest pixel-level AUROC for structural abnormalities;
3) $M_T$ is most effective for detecting logical abnormalities; 4) and unified detection of logical and structural anomalies requires integrating all three memory banks.
These results align with the original motivation for constructing multiple complementary memory banks in Sec.~\ref{section3_2}.

\begin{figure}[!t]
\centering
\includegraphics[width=\columnwidth]{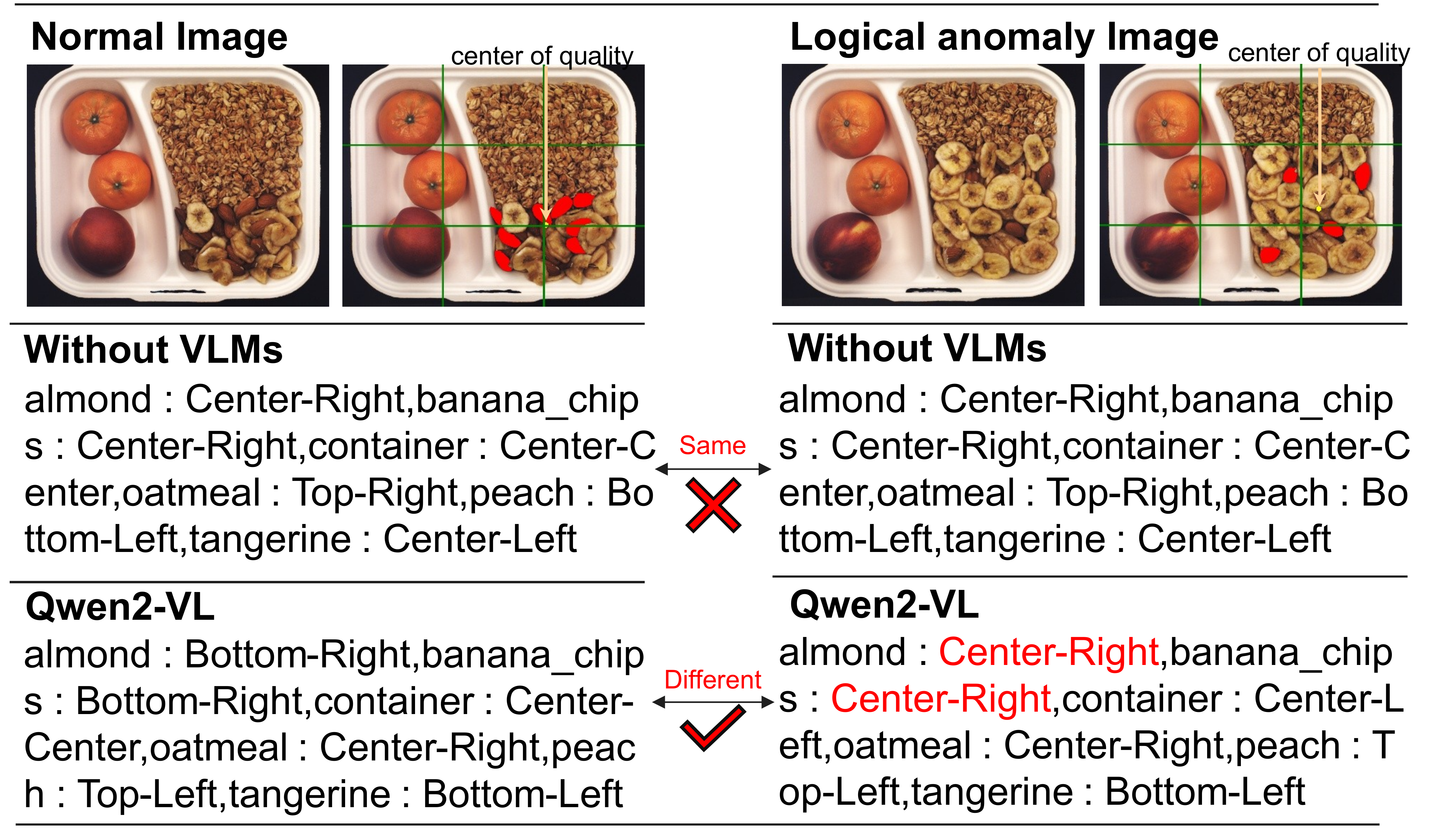} 
\caption{VLM's advantage in providing fuzzy locations for object categories facilitating logical anomaly detection on the MVTec LOCO \cite{bergmann2022beyond} dataset.}
\label{xy_VS_VLMs}
\end{figure}

\begin{table}[t]
\caption{Plug-and-play class-level text memory bank $M_T$ on the MVTec LOCO \cite{bergmann2022beyond} dataset.}
\centering
\small
\begin{tabular}{c|c}
\hline
Methods & AUROC \\ \hline
UniVAD \cite{gu2025univad} & 81.9 \\
UniVAD \cite{gu2025univad} + $M_{T}$ & 91.2 ($\uparrow$ 9.3)  \\
\hline
\end{tabular}
\label{method_with_text}
\vspace{-0.5cm}
\end{table}

\begin{figure}[!t]
\centering
\includegraphics[width=\columnwidth]{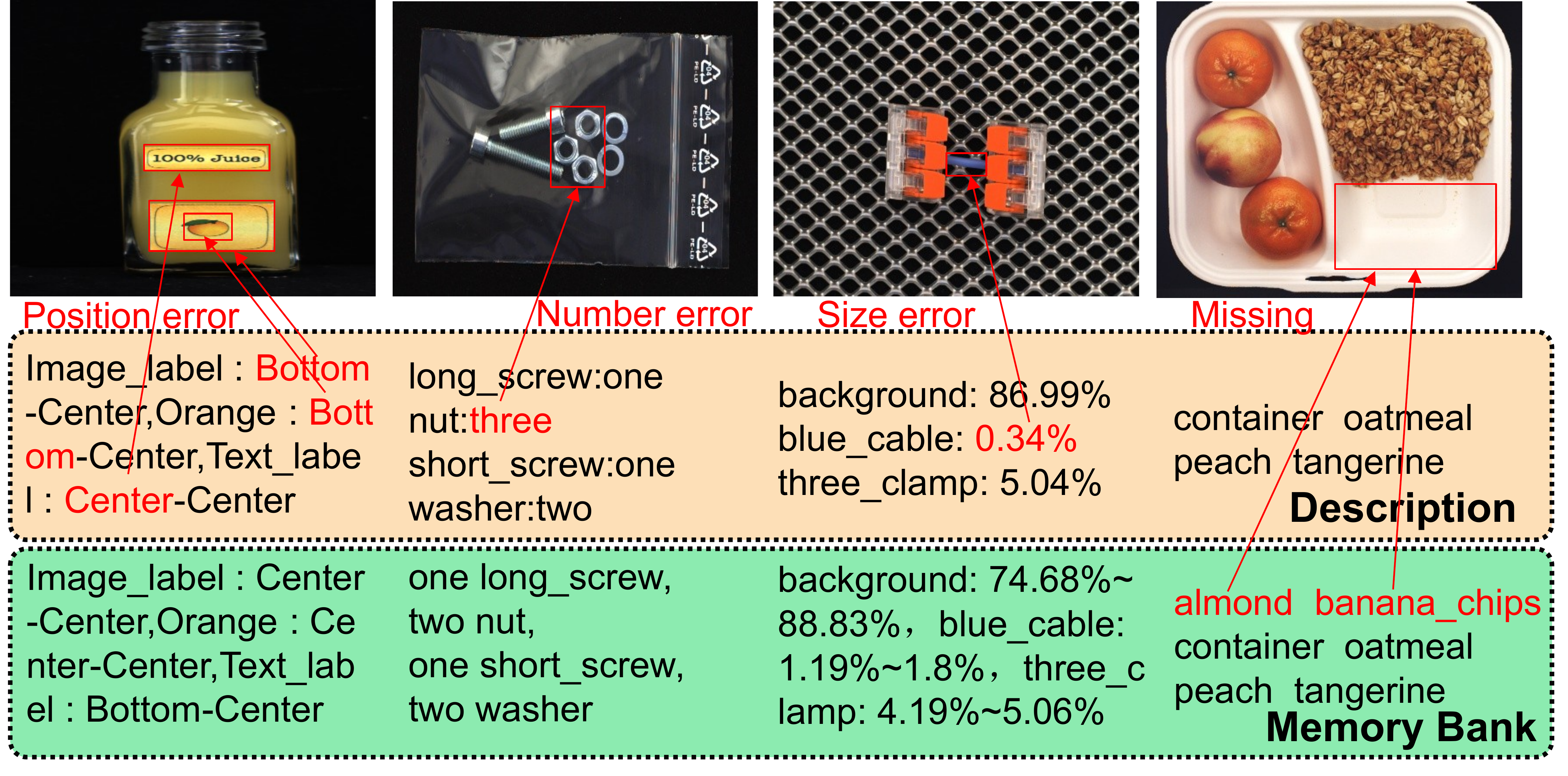} 
\caption{Logical anomaly detection by our text memory banks $M_T$ on the MVTec LOCO \cite{bergmann2022beyond} dataset.}
\label{text_breakfast}
\end{figure}

We conducted ablation studies on our class-level text memory bank $M_T$ using the MVTec LOCO dataset \cite{bergmann2022beyond}.
The image-level AUROC shown in Table~\ref{Text_Ablation} are calculated using $M_T$ only.
The first and second row represent directly inputting entire images into VLMs \cite{wang2024qwen2} to generate textual descriptions, with anomaly scores computed via cosine similarity between textual features encoded by the T5 encoder \cite{raffel2020exploring} and the CLIP text encoder \cite{radford2021learning}. Compared to the CLIP encoder, which primarily captures global text features of the entire image, the T5 encoder extracts richer semantic information, thereby improving logical anomaly detection.
The third row represents that images are first segmented with Grounded SAM \cite{ren2024grounded}, followed by VLM-based textual descriptions of each segmented object. These descriptions are concatenated and encoded with T5 encoder \cite{raffel2020exploring} to compute similarity and anomaly scores. This validates the necessity of first segmenting and then describing the image.
In the fourth row, the ``VLMs $\to O_{pos}^j \to T_{img}^j$" indicates that Algorithm~\ref{Algorithm1} is used to obtain class-level text descriptions. 
Using a T5 encoder to calculate text similarity and derive anomaly scores may lack sensitivity to the number, location, and size of objects, which are critical for logical anomaly detection. 
To address this limitation, 
our proposed class-level text matching strategy in Algorithm~\ref{Algorithm2} significantly enhance the performance of logical anomaly detection, as shown in the fifth row of Table~\ref{Text_Ablation}. In addition, we conducted an ablation study on the MVTec LOCO dataset \cite{bergmann2022beyond} to evaluate object localization using vision-language models (VLMs), as shown in Fig. \ref{xy_VS_VLMs}. Unlike the nine-grid method that estimates an object’s location by calculating its geometric center, VLMs leverage strong prior knowledge and semantic understanding to approximate object positions. This approach reduces errors that commonly arise when objects lie near equidistant boundaries.
Notably, our proposed class-level text memory bank can be seamlessly integrated into other methods in a plug-and-play manner, e.g., UniVAD \cite{gu2025univad} for unified logical and structural anomaly detection, PBAS \cite{10716437} for structural anomaly detection. Table \ref{method_with_text} demonstrate that incorporating our text memory bank substantially enhances the performance of logical anomaly detection. As shown in Fig.  \ref{text_breakfast}, we present the results of applying class-level textual memory bank to detect anomalies in various scenarios of the MVTec LOCO dataset \cite{bergmann2022beyond}.
The results demonstrate that our logic-aware textual descriptions effectively identify anomalies related to object location, number, size, missing, and other logical inconsistencies.

\begin{figure}[t]
\centering
\includegraphics[width=\columnwidth]{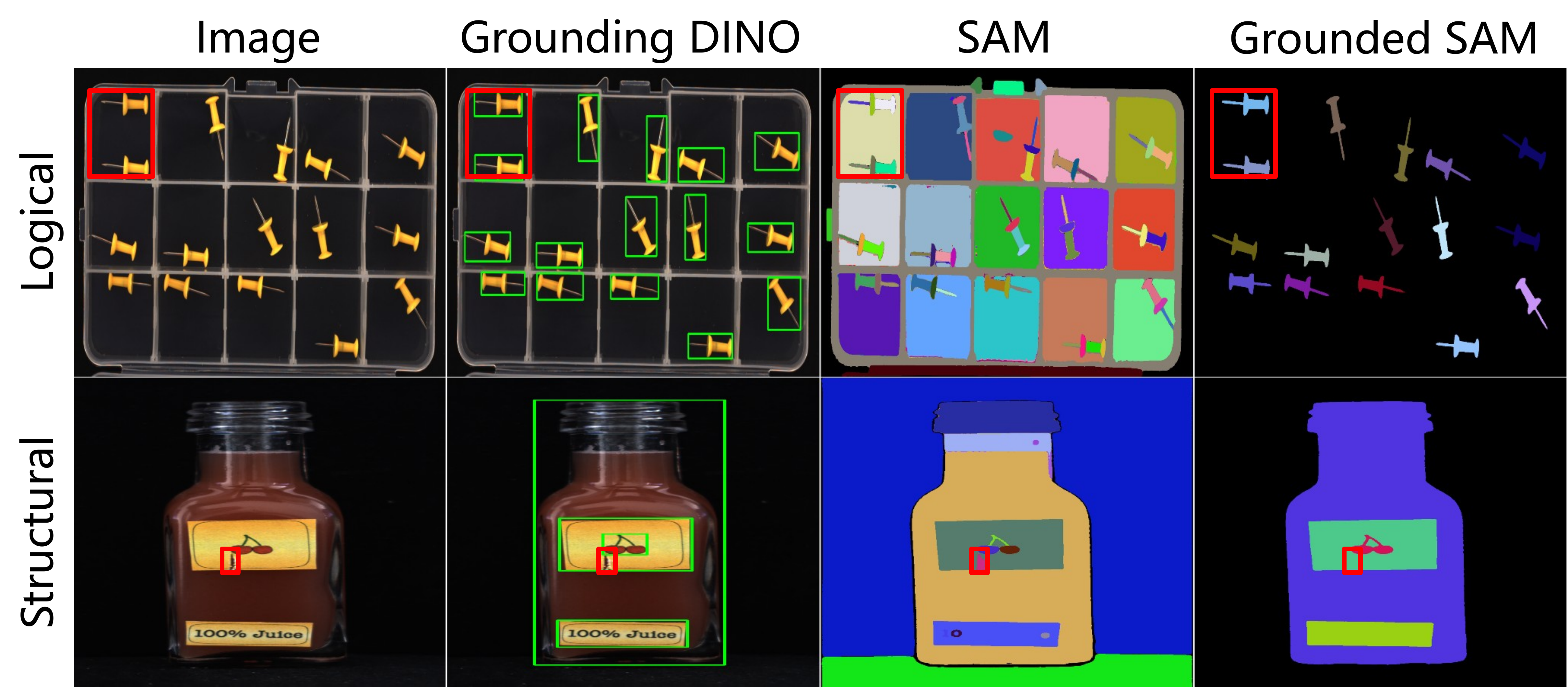} 
\caption{Visual comparison between SAM \cite{kirillov2023segment} and Grounding SAM \cite{ren2024grounded} on the MVTec LOCO \cite{bergmann2022beyond} dataset.}
\label{Segmentation_feature_map}
\end{figure}

We conducted ablation experiments on the segmentation model, as shown in Fig. \ref{Segmentation_feature_map}. The third and fourth columns present the results of SAM segmentation \cite{kirillov2023segment} and Grounding SAM segmentation \cite{ren2024grounded}, respectively.
For logical anomaly detection (first row), Grounding SAM \cite{ren2024grounded} produces semantically more complete segmented objects compared with SAM \cite{kirillov2023segment}. It effectively excludes most background regions unrelated to logical anomalies, enabling our TMUAD to focus more directly on the relevant objects.
For structural anomaly detection (second row), SAM \cite{kirillov2023segment}  outperforms Grounding SAM \cite{ren2024grounded} by capturing finer object details and comprehensively segmenting image regions that may contain anomalies. This detailed representation of SAM \cite{kirillov2023segment} is particularly beneficial for detecting structural anomalies.

\begin{table}[t]
\caption{Ablation studies of the object-level image memory bank $M_O$ on the MVTec LOCO \cite{bergmann2022beyond} dataset in image-level AUROC.}
\centering
\small
\begin{tabular}{c|c|c}
\hline
Encoder & Segmentation model & AUROC (only $M_O$) \\
\hline
DINOv2 \cite{oquab2023dinov2}& Grounded SAM \cite{ren2024grounded} & 65.8 \\
CLIP \cite{radford2021learning}& Grounded SAM \cite{ren2024grounded} & 67.7\\
DINOv2 \cite{oquab2023dinov2} & SAM only \cite{kirillov2023segment}  & \underline{74.9} \\
CLIP \cite{radford2021learning} & SAM only \cite{kirillov2023segment} & \textbf{75.1} \\
\hline
\end{tabular}
\label{label_MO}
\end{table}

\begin{table}[t]
\caption{Ablation studies of the patch-level image memory bank $M_P$ in image-level AUROC.}
\centering
\small
\resizebox{\columnwidth}{!}{
\begin{tabular}{c|c|c|c}
\hline
Dataset & DINOv2 \cite{oquab2023dinov2} & CLIP \cite{radford2021learning}& DINOv2+CLIP\\ \hline
MVTec LOCO \cite{bergmann2022beyond} & 74.8 & \underline{76.0} & \textbf{78.9} \\
MVTec AD \cite{bergmann2019mvtec} & 97.2 & \underline{98.7} & \textbf{98.9} \\
VisA \cite{zou2022spot} & 91.2 & \underline{94.3} & \textbf{95.5} \\ \hline
\end{tabular}
}
\label{label_DINO_CLIP}
\end{table}

\begin{figure}[!t]
\centering
\includegraphics[width=\columnwidth]{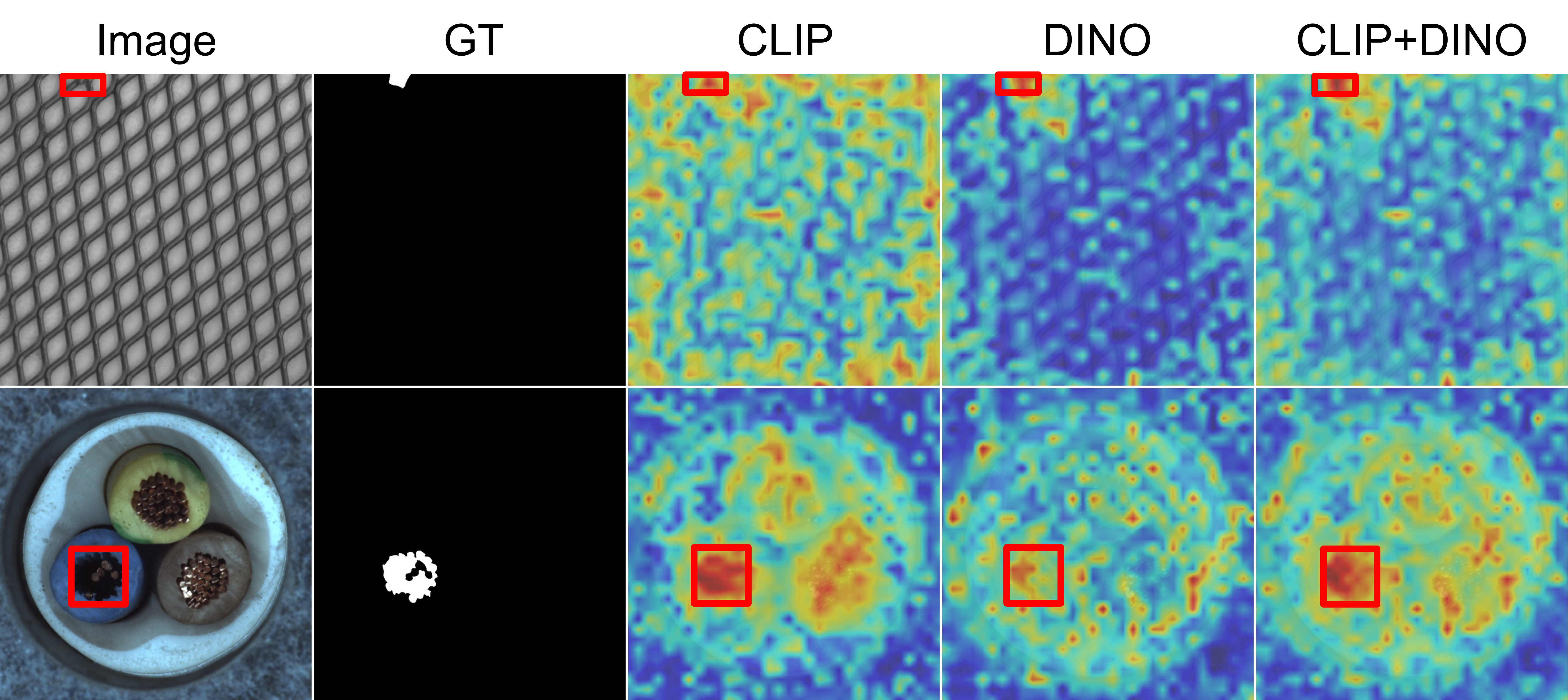} 
\caption{Visual results for ablation studies of the patch-level image memory bank $M_{P}$ on the MVTec AD \cite{bergmann2019mvtec} dataset.}
\label{dino-clip-encoder}
\end{figure}

We conducted ablation experiments on object-level image memory banks $M_O$ on the MVTec LOCO \cite{bergmann2022beyond} dataset. 
The image-level AUROC shown in Table~\ref{label_MO} are calculated using $M_O$ only.
Table \ref{label_MO} indicate that the optimal configuration involves using SAM \cite{kirillov2023segment} for segmentation, extracting object-level features with the CLIP image encoder \cite{radford2021learning}, computing text similarity, and deriving anomaly scores.

We conducted ablation experiments on the patch-level image memory bank $M_P$ on the MVTec LOCO \cite{bergmann2022beyond} dataset.
The image-level AUROC shown in Table~\ref{label_DINO_CLIP} are calculated using $M_P$ only.
Table \ref{label_DINO_CLIP} shows that combining DINOv2 \cite{oquab2023dinov2} and CLIP encoders \cite{radford2021learning} yields superior performance compared to using either encoder alone.
Fig. \ref{dino-clip-encoder} further illustrates their complementary strengths: CLIP emphasizes global differences in foreground objects, which is advantageous for detecting anomalies in large objects, whereas DINO captures fine-grained pixel-level differences, enabling more accurate detection of small-object anomalies.
This dual-encoder framework achieves effective complementarity in structural anomaly detection: 1) DINO suppresses noise in non-anomalous regions (first row), while 2) CLIP enhances the detection of large-object anomalies (second row).

\section{Discussion and Limitation}

\begin{figure}[!t]
\centering
\includegraphics[width=\columnwidth]{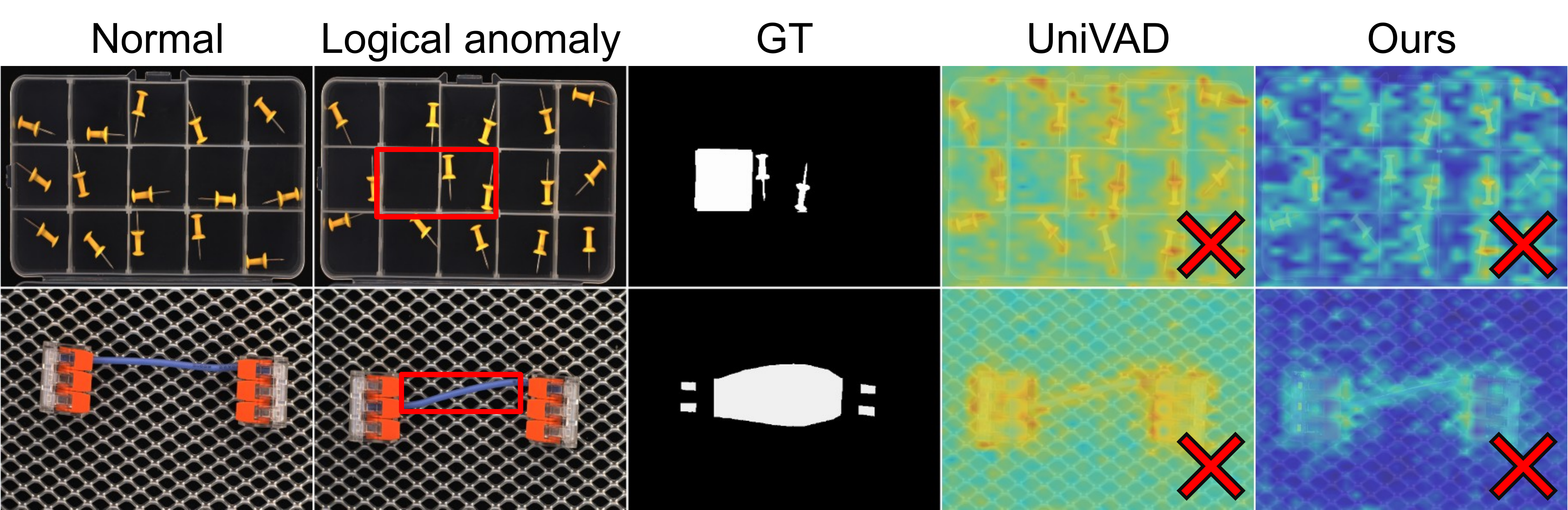} 
\caption{Failure cases of our method on the MVTec LOCO \cite{bergmann2022beyond} dataset.}
\label{challenge}
\end{figure}

\subsection{Limitations}
Our method sometimes fails to detect logical anomalies on the MVTec LOCO \cite{bergmann2022beyond} dataset, as shown in Fig. \ref{challenge}. In the first row, although the number of pins and cells is consistent, one cell contains two pins. This anomaly is challenging to identify because cells are generally treated as background; incorporating background anomaly detection could improve performance in such cases. In the second row, the wires are misaligned when connecting to the ports. This issue may be mitigated by incorporating the concept of object contact detection \cite{chen2023detecting,dwivedi2025interactvlm}.

\begin{table}[t]
\caption{Few shot (4-shot) image-level AUROC of unified anomaly detection on the MVTec LOCO \cite{bergmann2022beyond} dataset.}
\centering
\small
\resizebox{\columnwidth}{!}{
\begin{tabular}{c|c|c|c|c|c|c}
\hline
 \multirow{2}*{Methods} & PatchCore  &  WinCLIP+ & PromptAD & UniVAD & LogSAD & \multirow{2}*{Ours}\\ 
& \cite{roth2022towards} &  \cite{jeong2023winclip}  & \cite{10657290} & \cite{gu2025univad} &\cite{zhang2025towards} &\\ 
 \hline
AUROC (\%) & 68.7 & 71.3  & 73.5 & 76.0 & \textbf{86.3} & 81.2$\to$\underline{85.3}\\ 
\hline
\end{tabular}
}
\label{few_shot}
\end{table}

\begin{figure}[t]
\centering
\includegraphics[width=0.95\columnwidth]{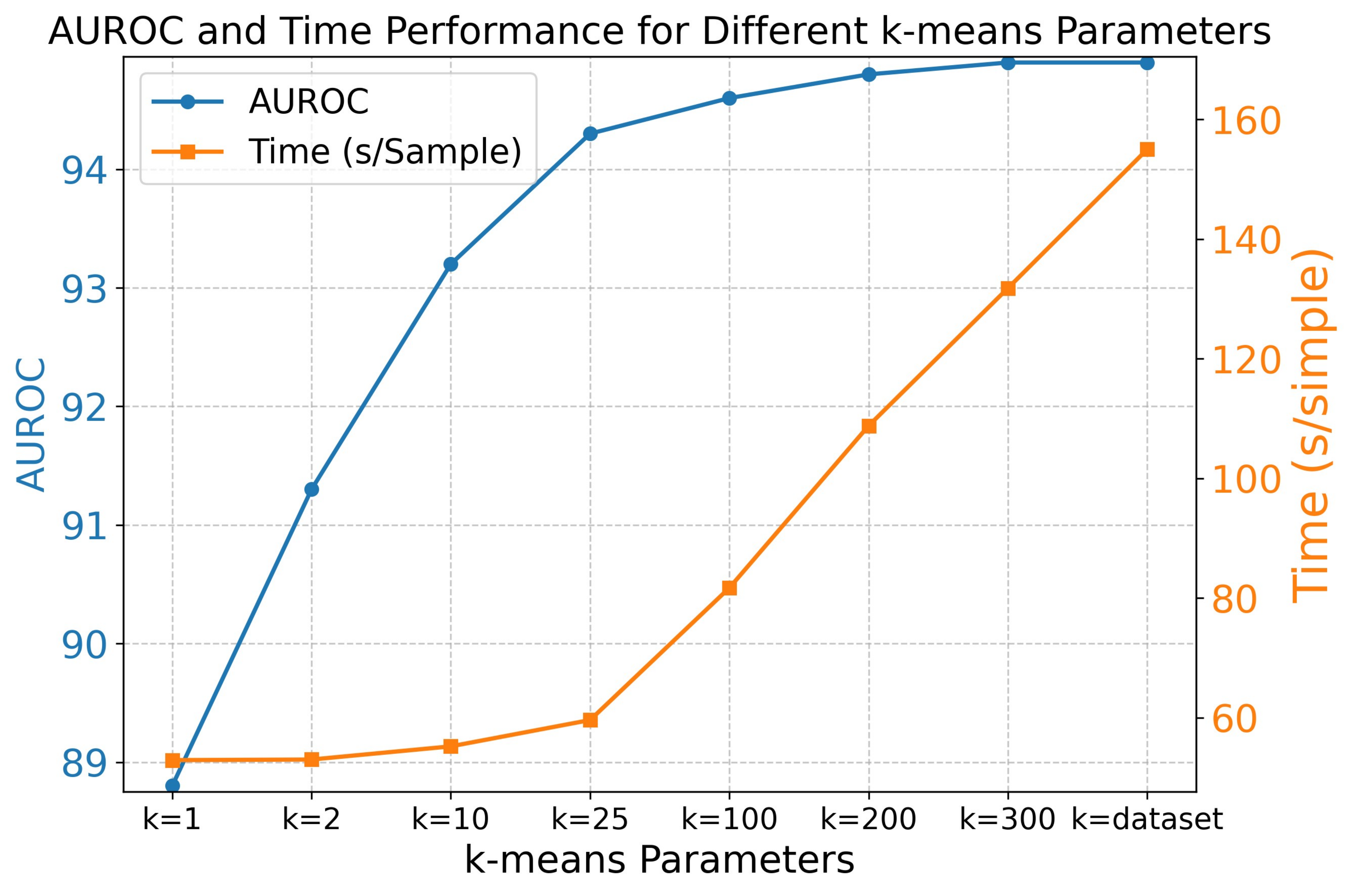} 
\caption{Ablation study of the number of clusters in K-means of $M_P$ on the MVTec LOCO \cite{bergmann2022beyond} in image-level AUROC.}
\label{kmeans_performance_comparison}
\end{figure}

\begin{table}[t]
\caption{Resource efficiency and performance analysis on the MVTec LOCO \cite{bergmann2022beyond} dataset in image-level AUROC.
``-I" indicates CLIP TRAINED INDIVIDUALLY on the MVTec LOCO \cite{bergmann2022beyond} dataset, while ``-U" denotes CLIP TRAINED uniformly across all datasets.
}
\centering
\small
\resizebox{\columnwidth}{!}{
\begin{tabular}{c|c|c|c}
\hline
Methods & Inference Time (s) & Parameters (M)  & AUROC\\ \hline
DeCo-Diff \cite{beizaee2025correcting} & \underline{29.0} & \underline{478.6}  & 69.5\\ 
PBAS \cite{10716437} & \textbf{0.9} & \textbf{72.8}  & 81.4\\ 
UniVAD \cite{gu2025univad} & 1286.1 & 2402.1 & 83.3\\ 

\hline
TMUAD-U-L & 131.8 & 11741.0  & \underline{94.9}\\ 
TMUAD-U-B ({\bf Ours}) & 81.7 & 11741.0  & 94.6 \\ 
TMUAD-U-B (3 GPUs) & 30.7 & 11741.0  & 94.6 \\
TMUAD-U-S & 32.9 & 3449 & 94.1 \\
\hline
TMUAD-I-B & 81.7 & 11741.0 & \textbf{95.2} \\
PBAS-I-B & 33.9 & 9534.8 & 92.0\\ 
\hline
\end{tabular}
}
\label{run_time}
\end{table}

\subsection{Few-shot Anomaly Detection}
We evaluated few-shot anomaly detection on the MVTec LOCO \cite{bergmann2022beyond} dataset, as shown in Table \ref{few_shot}. The results indicate that our TMUAD (AUROC: 81.2) performs worse than LogSAD \cite{zhang2025towards} (AUROC: 86.3) in few-shot settings but TMUAD (AUROC: 94.6) outperforms LogSAD \cite{zhang2025towards} (AUROC: 90.2) in full data (Table \ref{mvtec_loco}). 
In addition, our TMUAD with a relaxed class-level text matching strategy  (i.e., w/o $O_{num}^j,O_{pos}^j,O_{size}^j$) (AUROC: 85.3) demonstrates competitive performance with LogSAD \cite{zhang2025towards} (AUROC: 86.3) in few-shot scenarios.
We analyze that the reason may be that our strict text matching strategy is highly effective when sufficient normal samples are available, but overly restrictive in settings with limited data, leading to performance degradation in few-shot scenarios.

\subsection{Resource Efficiency Analysis}

Fig. \ref{kmeans_performance_comparison} shows that: 1) the AUROC metric improves as the number of clusters increases, but the gain saturates around 100 clusters; and 2) when the number of clusters exceeds 100, the average inference time rises sharply. 
Table \ref{run_time} shows that performance and efficiency comparisons between our TMUAD under various settings and the SOTA methods.
We constructed our base model, TMUAD-U-B, using Qwen2-VL-7B with 100 K-means clusters in the memory bank $M_P$. We further introduce two variants: TMUAD-U-L, built on Qwen2-VL-7B with 300 clusters, and TMUAD-U-S, built on w/o Qwen2-VL-7B and with 25 clusters. 
Additionally, we report the inference time of TMUAD-U-B when accelerated  by three A6000 GPUs in parallel. Except TMUAD-U-B (3 GPUs), all experiments in this paper are conducted on a single A6000 GPU.
In future work, we will accelerate anomaly detection by applying product quantization \cite{jegou2010product} during the search for similar samples.

\section{Conclusion}
The proposed TMUAD is a memory-based anomaly detection framework that unifies logical and structural anomaly detection by leveraging multiple memory banks, consisting of class-level text, object-level image, and patch-level image memory bank. Our class-level text memory bank enhances logical anomaly detection by leveraging logic-aware textual descriptions, while our image memory bank is sensitive to structural anomalies through object-level features that preserve contours and patch-level features that capture global contextual information. Our text memory bank is plug-and-play for structural anomaly detection methods, enabling their extension into a unified framework for both logical and structural anomaly detection.
Our future work will focus on accelerating visual anomaly detection models and deploying them to robots, designing more effective feature representations, and incorporating object contact detection to improve anomaly detection.



%

\bibliographystyle{ieeetr}
\bibliography{reference}











\begin{IEEEbiography}[{\includegraphics[width=1in,height=1.25in,clip,keepaspectratio]{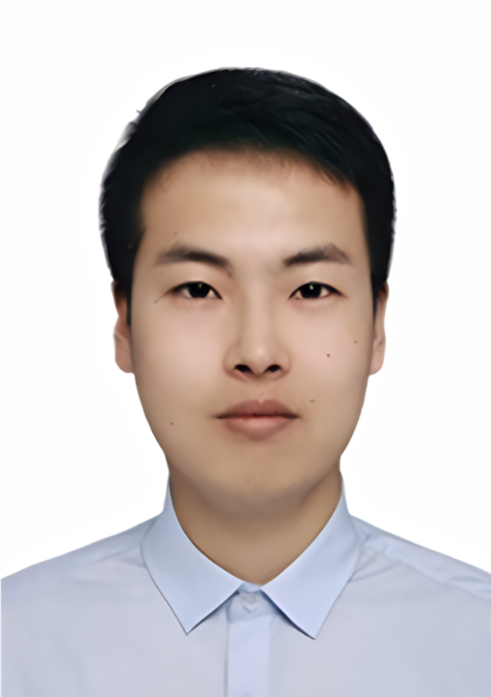}}]{Jiawei Liu}
   (Member, IEEE) received his B.E. degree from Northeast Agricultural University, Harbin, China, in 2018, and his Ph.D. degree in Pattern Recognition and Intelligent Systems from the Shenyang Institute of Automation, Chinese Academy of Sciences, Shenyang, China, in 2024. 
   He is currently a research assistant professor at Shenyang Institute of Automation, Chinese Academy of Sciences. His research interests include deep learning, illumination processing, image restoration, shadow removal, anomaly detection, and diffusion models.
   \end{IEEEbiography}
   \begin{IEEEbiography}[{\includegraphics[width=1in,height=1.25in,clip,keepaspectratio]{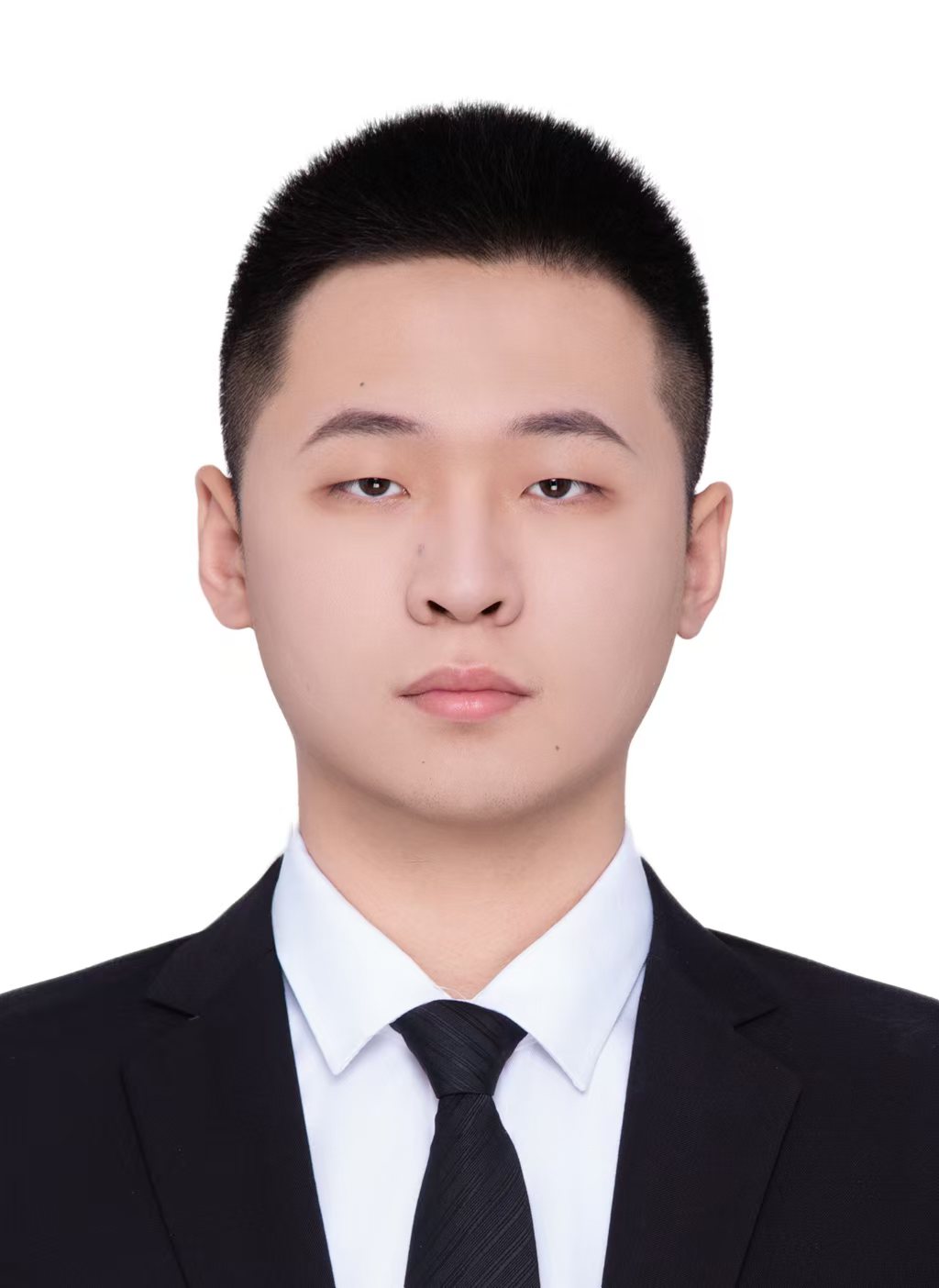}}]{Jiahe Hou} received his B.E. degree from Dalian University of Technology in 2025. 
      Currently, he is a Ph.D. student at Shenyang Institute of Automation, Chinese Academy of Sciences. His current research interests include anomaly detection and vision language models.
      \end{IEEEbiography}
\begin{IEEEbiography}[{\includegraphics[width=1in,height=1.25in,clip,keepaspectratio]{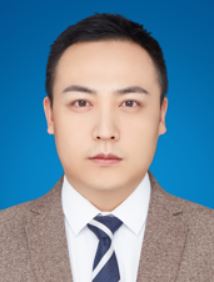}}]{Wei Wang} received his B.E. degree and M.E. degree from Xidian University, Xi’an, China, In 2014, he received the Ph.D. degree from Shenyang Institute of Automation, Chinese Academy of Sciences, China. 
   He is currently a professor at the Shenyang Institute of Automation, Chinese Academy of Sciences, and deputy director of the Intelligent Inspection and Equipment Research Laboratory.
   \end{IEEEbiography}
   \begin{IEEEbiography}[{\includegraphics[width=1in,height=1.25in,clip,keepaspectratio]{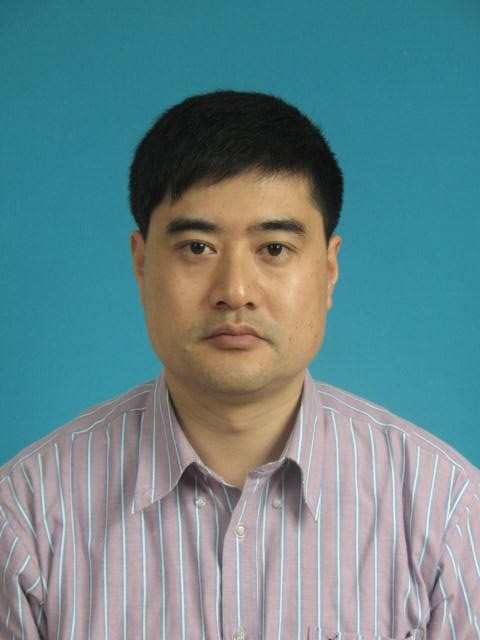}}]{Jinsong Du} received his B.E. degree from Shenyang University of Technology, Shenyang, China, in 1993, and his Ph.D. degree in Chinese Academy of Sciences (CAS) Graduate School, Chinese Academy of Sciences, Shenyang, China, in 2010. 
   He is currently a professor at the Shenyang Institute of Automation, Chinese Academy of Sciences, where he serves as director of the Intelligent Inspection and Equipment Research Laboratory.
   \end{IEEEbiography}
   \begin{IEEEbiography}[{\includegraphics[width=1in,height=1.25in,clip,keepaspectratio]{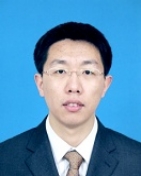}}]{Yang Cong} (Senior Member, IEEE) received the B.S. degree from Northeast University in 2004 and the Ph.D. degree from the State Key Laboratory of Robotics, Chinese Academy of Sciences, in 2009. From 2009 to 2011, he was a Research Fellow with the National University of Singapore (NUS) and Nanyang Technological University (NTU). He was a Visiting Scholar with the University of Rochester. He was the professor until 2023 with Shenyang Institute of Automation, Chinese Academy of Sciences. He is currently the full professor with South China University of Technology. He has authored over 80 technical articles. His current research interests include robot, computer vision, machine learning, multimedia, medical imaging and data mining. He has served on the editorial board of the several joural papers. He was a senior member of IEEE since 2015. 
   \end{IEEEbiography}
   \begin{IEEEbiography}[{\includegraphics[width=1in,height=1.25in,clip,keepaspectratio]{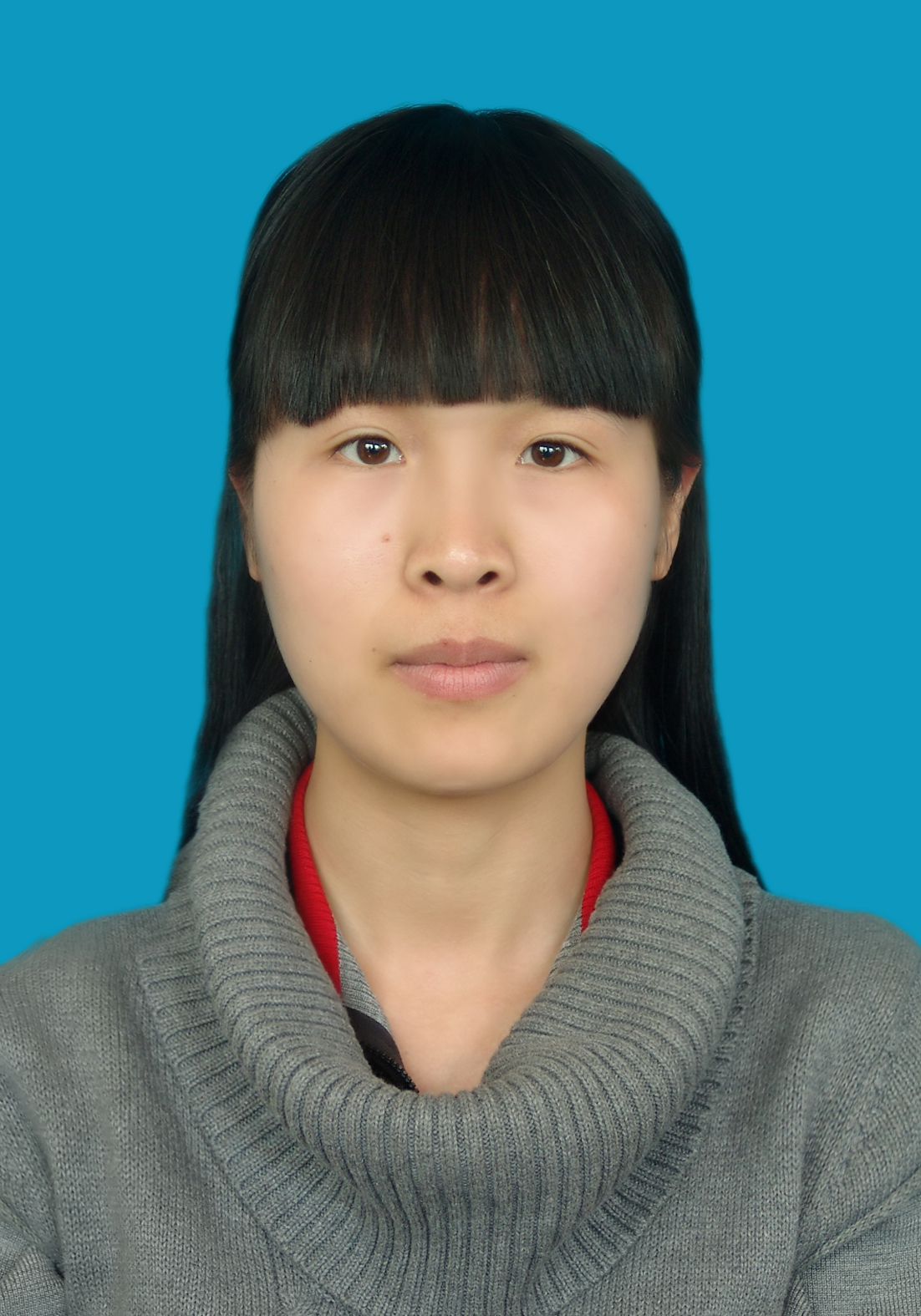}}]{Huijie Fan} (Member, IEEE) received the B.E. degree in automation	from the University of Science and Technology of Science and Technology of China, China, in 2007, and the Ph.D. degree in mode recognition and intelligent systems from the Chinese Academy of Sciences University, Beijing, China, in 2014. She is currently a professor with the Institute of Shenyang Automation of the Chinese Academy of Sciences. Her research interests include deep learning on image processing and medical image processing and applications.
   \end{IEEEbiography}
\newpage

\vfill

\end{document}